%% file: main_ref.tex
\documentclass{ieeeaccess}
\usepackage[utf8]{inputenc}
\usepackage[english]{babel}
\usepackage{multirow}
\usepackage[dvipsnames*,svgnames,table]{xcolor}
\definecolor{accessblue}{RGB}{0,105,154}
\definecolor{greycolor}{cmyk}{0,0,0,1}
\usepackage{textcomp}
\usepackage{graphicx}
\usepackage{lipsum} 
\usepackage{hyperref,color,breqn,multirow}
\usepackage[top=0.7in, left=0.65in]{geometry}
\usepackage{amsmath,amsfonts,amssymb,amscd,bm}
\usepackage{algorithm}
\usepackage{algpseudocode}
\usepackage{csquotes}
\usepackage{booktabs}
\usepackage{tikz}
\usepackage{tabularx}
\usepackage{array}
\usepackage{hyperref,color,breqn,multirow}
\usepackage[top=0.7in, left=0.65in]{geometry}
\usepackage{graphicx}
\usepackage{cleveref}
\usepackage{subcaption}
\usepackage{upgreek}
\NewSpotColorSpace{PANTONE}
  \AddSpotColor{PANTONE} {PANTONE3015C} {PANTONE\SpotSpace 3015\SpotSpace C} {1 0.3 0 0.2}
  \SetPageColorSpace{PANTONE}\usetikzlibrary{shapes,shapes.geometric,arrows,positioning,patterns,shapes.arrows,fit}
\usepackage{tuda-pgfplots}
\usepackage{pgfplots}
\DeclareUnicodeCharacter{2212}{−}
\usepgfplotslibrary{groupplots,dateplot}
\usepgfplotslibrary{groupplots}
\usetikzlibrary{patterns,shapes.arrows}
\pgfplotsset{compat=newest}
\usepackage[normalem]{ulem}
\usepackage[labelfont=it,textfont={bf}]{caption}
\DeclareCaptionFormat{citation}{%
   \ifx\captioncitation\relax\relax\else
     \captioncitation\par
   \fi
   #1#2#3\par}

\let\captioncitation\relax
\usepackage[labelfont=bf,font=footnotesize,textfont=scriptsize,singlelinecheck=off,justification=centering]{subcaption}
\makeatletter
\newcommand*{\rom}[1]{\expandafter\@slowromancap\romannumeral #1@}

\makeatletter
\newcommand{\removelatexerror}{\let\@latex@error\@gobble}
\makeatother

\usepackage[style=ieee]{biblatex}
\usepackage{tabularx}
\usepackage{afterpage}
\usepackage{adjustbox}
\usepackage{animate} 
\usepackage[
singlelinecheck=false 
]{caption}
\usepackage{hyperref}
\addto\extrasenglish{%
}
\makeatother

\newcommand{\kpi}{k}
\allowdisplaybreaks
\makeatletter
\newcommand\notsotiny{\@setfontsize\notsotiny{6}{7}}
\makeatother
\usepackage{eso-pic}
\AddToShipoutPicture*{\footnotesize\sffamily\raisebox{1cm}{\hspace{1.65cm}\fbox{\parbox{\textwidth}{\copyright 2023 IEEE.  Personal use of this material is permitted.  Permission from IEEE must be obtained for all other uses, in any current or future media, including reprinting/republishing this material for advertising or promotional purposes, creating new collective works, for resale or redistribution to servers or lists, or reuse of any copyrighted component of this work in other works.}}}}

\begin{document}
\history{Date of publication xxxx 00, 0000, date of current version xxxx 00, 0000.}
\doi{10.1109/ACCESS.2023.DOI}

\title{Deep learning based Meta-modeling for Multi-objective Technology Optimization of Electrical Machines}
\author{
\uppercase{Vivek Parekh}\authorrefmark{1,2}, 
\uppercase{Dominik Flore}\authorrefmark{2}, 
\uppercase{Sebastian Sch\"{o}ps}\authorrefmark{1}
}

\address[1]{Computational Electromagnetics Group,Technische Universit\"{a}t Darmstadt,64289, Germany }
\address[2]{Powertrain Solutions, Mechanical Engineering and Reliability,Robert Bosch GmbH,70442 Stuttgart, Germany}

\tfootnote{This work is financially supported by Robert Bosch GmbH.}

\markboth
{V. Parekh \headeretal: Deep learning based Meta-modeling for Multi-objective Technology Optimization of Electrical Machines}
{V. Parekh \headeretal: Deep learning based Meta-modeling for Multi-objective Technology Optimization of Electrical Machines}

\corresp{Corresponding author: Vivek Parekh (e-mail: Vivek.Parekh@de.bosch.com).}

\begin{abstract} 
Optimization of rotating electrical machines is both time- and computationally expensive. Because of the different parametrization, design optimization is commonly executed separately for each machine technology. In this paper, we present the application of a variational auto-encoder (VAE) to optimize two different machine technologies simultaneously, namely an asynchronous machine and a permanent magnet synchronous machine. After training, we employ a deep neural network and a decoder as meta-models to predict global key performance indicators (KPIs) and generate associated new designs, respectively, through unified latent space in the optimization loop. Numerical results demonstrate concurrent parametric multi-objective technology optimization in the high-dimensional design space. The VAE-based approach is quantitatively compared to a classical deep learning-based direct approach for KPIs prediction.
\end{abstract}


\titlepgskip=-15pt
\maketitle

\section{Introduction}\label{sec:INT}

\PARstart{E}{lectrical} machines play a pivotal role in the modern era, powering everything from home appliances to electric vehicles and industrial equipment. To reduce manufacturing costs, electrical machines are numerically optimized via virtual prototyping, which involves finite element (FE) simulations or analytical calculations before the actual machine is constructed.
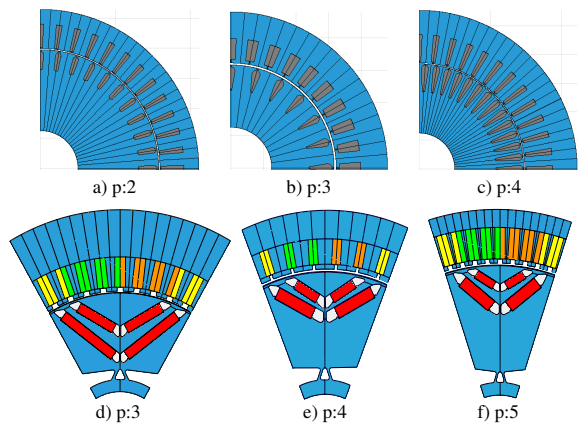
\begin{figure}[!htp]
	\centering
	\input{fig/data_visualization}
	\caption{Representative geometries of ASM (a-c) and PMSM (d-f) for different pole pairs (p).}
	\label{fig:dvs}
\end{figure}
Design evaluations using these classical techniques are both costly and time-consuming. Therefore, it is essential to find faster optimization methods to ensure a more sustainable and energy-efficient design workflow. 

In recent years, there has been a significant increase in the use of machine learning-based meta-modeling for the accelerated numerical optimization of electrical machines for various purposes. For example, the study in \cite{8661767} shows how trained data-driven deep learning (DL) models estimate magnetic field distribution for different low-frequency electromagnetic devices such as a transformer, a coil in air, and an interior permanent magnet machine. In another work \cite{9829204}, a convolutional neural network (CNN), a type of a deep neural network (DNN), is used as a meta-model for diagnosing stator winding faults of a permanent magnet synchronous machine (PMSM). The paper \cite{9291033} presents how various machine learning methods assist in fault detection for induction machines. The concurrent application of variational autoencoder (VAE) and DNN for technology optimization of electromagnetic devices is investigated in \cite{electronics10182185}. An unsupervised learning-based anomaly detection model using a VAE for fault diagnosis in electric drive systems is presented in \cite{9773565}. Several successful works of machine learning-based meta-modeling for the optimization of electromagnetic devices have been discussed in \cite{9745918}. Many articles\cite{8056321,8785344,9940692,9910710,8921886,9982881,app13031395} demonstrate the successful application of different machine-learning approaches at different stages of the design and optimization of electrical machines. The reduction of computational time required to generate a sufficient amount of data for training large-scale machine learning-based meta-models needs to be addressed. In order to tackle this issue, a method is proposed in \cite{9896140} for generating a large amount of data from a small number of FE simulation results using a deep generative model and a CNN. In a recent study \cite{10036443}, an approach for topology optimization of PM motors using the VAE and the neural network was demonstrated to generate various shapes and predict their corresponding motor characteristics within the optimization loop. However, the VAE often fails while reconstructing images in this approach. In \cite{9333549}, it has been shown how cross-domain key performance indicators (KPIs) can be estimated with high accuracy for different input representations of PMSM, i.e., image-based and parametric input using various DNNs. The parameter-based representation was observed to be more suitable concerning prediction accuracy with less computational effort than the image-based model. However, it can not deal with multiple topologies of PMSM concurrently. In \cite{9745548}, it is demonstrated how we can circumvent this problem for differently parameterized topologies of PMSM using VAE. The encoder maps the complex, high-dimensional combined input design space into a lower-dimensional unified latent distribution. In the latent space, multi-topology objective optimization was performed by predicting KPIs using DNN and generating associated new designs with the decoder. 

So far, to the best of our knowledge, machine learning methods have been applied for a scenario dealing with the numerical optimization of a single machine type at a time, for example, PMSM, asynchronous machine (ASM), or DC machine. In this paper, we aim to perform concurrent multi-technology objective optimization (MTOO) for two different machine types, namely PMSM and ASM. Both machine types are distinctly parameterized and operate on different working principles. This difference can affect the performance and efficiency of the machines in various applications. For example, PMSMs typically have higher efficiency and power density than ASMs but may also be more expensive to manufacture due to the cost of the permanent magnets. 

This paper proposes two significant differences compared to our previous works. We consider, in addition to geometric parameters, more challenging varying topological parameters such as pole pairs, number of slots per pole per phase, winding connection (star or delta), etc. Secondly, in order to handle these challenging parameters and the additional zeros in the combined design space, we propose a new optimization procedure that enhances synchronization between the decoder and the KPI predictor in the latent space. This proposed procedure improves the prediction accuracy of KPIs and associated design parameters. Additionally, we provide numerical analysis of the direct DL approach using a DNN for KPIs prediction\cite{9333549}.

The paper is organized as follows: in the next section, we explain the dataset details and training workflow. \Cref{sec:NAT} discusses the network architecture and training details. Quantitative results are demonstrated in \Cref{sec:NR}, and finally, the work is concluded in \Cref{sec:conclusion}.

\section{Dataset, training procedure and MOO}\label{sec:DM}
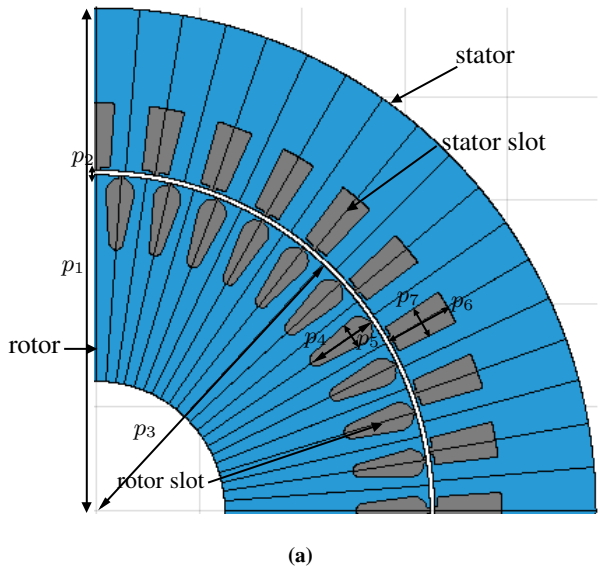
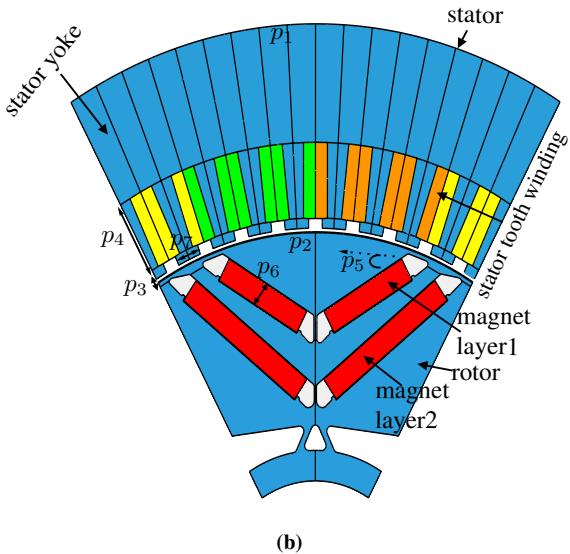
\begin{figure}
\centering
\begin{subfigure}[b]{0.5\textwidth}
  \centering
 	\input{fig/asm_technology_sample}
 	\vspace{-3mm}
    \caption{}
    \label{fig:TOP_ASM} 
\end{subfigure}

\begin{subfigure}[b]{0.5\textwidth}
   \centering
 	\input{fig/psm_technology_sample1}
 	\vspace{-3mm}
    \caption{}
    \label{fig:TOP_PSM}
\end{subfigure}
\caption[]{Different machine technologies (a) ASM (b) PMSM.}
\end{figure}
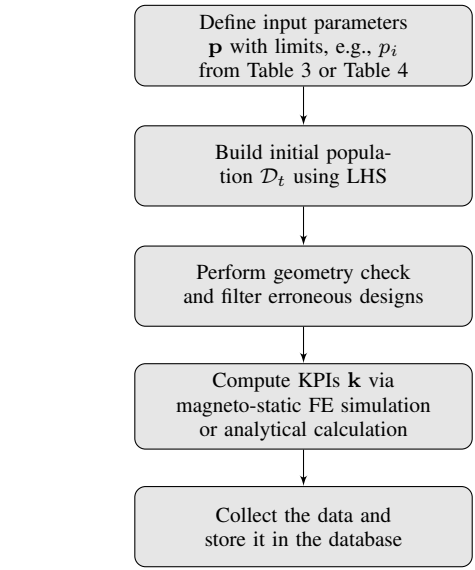
\begin{figure}[h!]
	\centering
	\input{fig/flow_chart_dataset_generation}
	\caption{General workflow for electrical machine data-generation.}
	\label{fig:dg}
\end{figure}
 This section is divided into three subsections: the  datasets detail, for explaining the VAE-based training workflow, and the last for formulating the multi-objective optimization (MOO) problem. 

\subsection{Dataset}
For the study, we build two datasets: one for an ASM and one for a PMSM. The usual industrial workflow of data generation is explained in \autoref{fig:dg} for any electrical machine class. There may be some modifications in the workflow at the phase when the design evaluation is conducted. In this study, we evaluate a PMSM design with a time-intensive magneto-static FE simulation (see \cite{Salon_1995aa}), while the ASM design is evaluated with analytical calculations (see \cite{Gerling2015-4,boldea2018induction}). The initial steps are identical in both instances of data generation. For example, specifying design parameters with constraints, creating a population with a Latin hypercube sampling technique \cite{mckay2000comparison} to cover the entire design space, and doing geometry checks for filtering erroneous designs using computer-aided design software (e.g.,\cite{geuzaine2009gmsh}).
\begin{table}[!h]
	\caption{KPIs information}
	\label{tab:kpis_info}
	\newcolumntype{Y}{>{\centering\arraybackslash}X}
	\begin{tabularx}{\linewidth}{|Y|X|Y|Y|Y|}
		\hline
        & \textbf{KPIs}              & \multicolumn{1}{c|}{\textbf{Value}}\\ \hline
		$\kpi_{1}$  & Material cost              & \multicolumn{1}{c|}{Euro}\\ \hline
		$\kpi_{2}$  & Maximum power              & \multicolumn{1}{c|}{kW}  \\ \hline
		$\kpi_{3}$  & Maximum torque             & \multicolumn{1}{c|}{Nm}\\ \hline
		\end{tabularx}%
\end{table}

\begin{table}[!h]
	\caption{System parameters}
	\label{tab:bcond}
	\begin{tabularx}{\linewidth}{|c|X|r|r|c|}
		\hline
        & \textbf{System parameter}                   & \multicolumn{2}{c|}{\textbf{Value}} & \textbf{Unit} \\ \hline
		$c_{1}$  & Inverter input DC voltage          & \multicolumn{2}{c|}{650}        & V         \\ \hline
		$c_{2}$  & Inverter input DC current          & \multicolumn{2}{c|}{400}          & A         \\ \hline
		$c_{3}$  & Rotational speed                   & \multicolumn{2}{c|}{$[1, 16000]$ (step size: 1000)}  & rpm \\ \hline
	\end{tabularx}%
\end{table}

\subsubsection{Dataset: ASM}

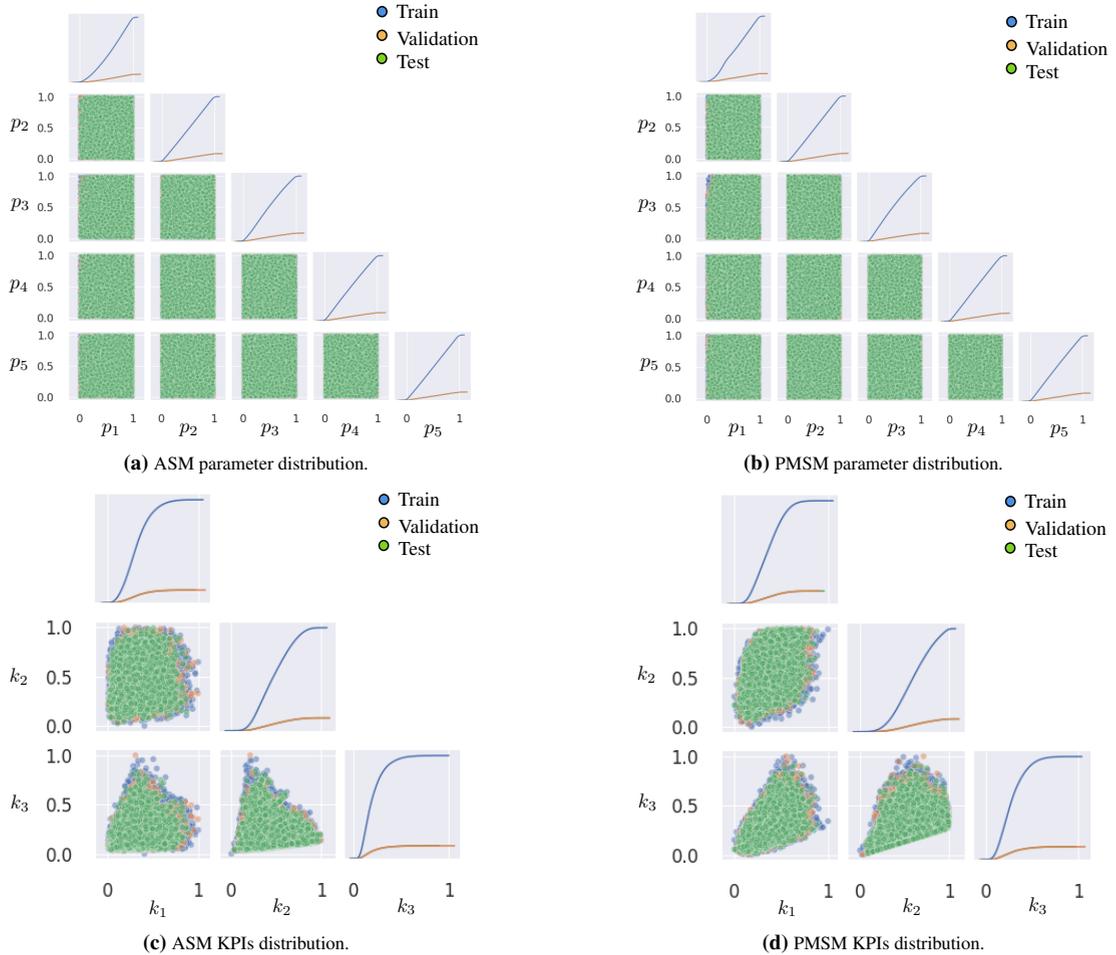
\begin{figure*}[!h]
    \centering
    \begin{subfigure}{0.49\linewidth}
        \centering
        \input{fig/asm_param_distr}
        \caption{ASM parameter distribution.}
        \label{fig:PDS1}
    \end{subfigure}
    \hfill
    \begin{subfigure}{0.49\linewidth}
        \centering
        \input{fig/psm_param_distr}
        \caption{PMSM parameter distribution.}
        \label{fig:PDS2}
    \end{subfigure}
    \\
    \begin{subfigure}{0.49\linewidth}
        \centering
        \input{fig/asm_kpi_distr}
        \caption{ASM KPIs distribution.}
        \label{fig:KPIDS1}
    \end{subfigure}
    \hfill
    \begin{subfigure}{0.49\linewidth}
        \centering
        \input{fig/psm_kpi_distr}
        \caption{PMSM KPIs distribution.}
        \label{fig:KPIDS2}
    \end{subfigure}
    \caption{Visualization parameter and KPIs distribution.}
    \label{fig:kdr}
\end{figure*}

\begin{table}
	\caption{ASM design parameters}
	\label{tab:input_asm}
	\begin{tabularx}{\linewidth}{|c|X|r|r|c|}
		\hline
		         & \textbf{Input design parameter}                 & \textbf{Min} & \textbf{Max}     & \textbf{Unit} \\ \hline
		$p_{1}$  & Stator outer diameter            &   159        &    232       & mm            \\ \hline
		$p_{2}$  & Air gap                          &   0.65       &   1.7        & mm            \\ \hline
		$p_{3}$  & Rotor outer diameter             &   85         &   190        & mm            \\ \hline
		$p_{4}$  & Rotor slot height                &   10         &   21         & mm            \\ \hline
		$p_{5}$  & Rotor slot width                 & 0.6          &  1.5         & mm            \\ \hline
		\hline
		         & \textbf{Discrete parameters}       & \multicolumn{2}{c|}{\textbf{Value}} & \textbf{Unit} \\ 
		\hline
		$p_{15}$  & Slots per pole per phase         & \multicolumn{2}{c|}{[2-4]}          & -         \\ \hline
		$p_{16}$  & Pole pairs (p)                   & \multicolumn{2}{c|}{[2-4]}          & -          \\ \hline
        $p_{17}$  & Stator winding connection        & \multicolumn{2}{c|}{star/delta}     & -          \\ \hline
        $p_{18}$  & Winding scheme                   & \multicolumn{2}{c|}{short pitch/full pitch}  & -  \\ \hline
		\hline

	\end{tabularx}%
\end{table}
We selected $T_{\textrm{ASM}}:= 50387$ valid ASM designs from the initial population. There is no fixed number of samples for data generation, but a large number of samples is usually preferred for the data-driven DL approach. There are total $d_{1}:= 18$ varying design parameters chosen for this data generation. From all the varying parameters, a few essential design parameters are detailed in \autoref{tab:input_asm}. Some parameters from \autoref{tab:input_asm} are shown in \autoref{fig:TOP_ASM}. Representative samples with varying pole pairs can be seen in \autoref{fig:dvs}. \autoref{fig:PDS1} and \autoref{fig:KPIDS1} visualize distribution of the listed parameters and KPIs. The dataset contains topology changing parameters such as the number of slots per pole per phase $p_{15}$, and varying pole pairs $p_{16}$. Electrical parameters are included as design parameters, i.e., stator winding connection (star/delta) and winding scheme (short pitch/full pitch winding). The evaluation time for one design is about $5-7$ minutes on a single-core CPU.  

\subsubsection{Dataset:PMSM}

\begin{table}
	\caption{PMSM design parameters}
	\label{tab:input_psm}
	\begin{tabularx}{\linewidth}{|c|X|r|r|c|}
		\hline
		         & \textbf{Input design parameter}  &\textbf{Min} & \textbf{Max}     & \textbf{Unit}  \\ \hline 
		$p_{1}$  & Stator outer diameter            &   159       &    232           & mm             \\ \hline
		$p_{2}$  & Rotor outer diameter             &   100       &    197           & mm              \\ \hline
		$p_{3}$  & Air gap                          &   0.8       &    2.2           & mm             \\ \hline
		$p_{4}$  & Stator tooth height              &   10        &   20             & mm             \\ \hline
		$p_{5}$  & Angle of magnet layer1           &  17         &   32             & mm             \\ \hline
		
        \hline
		         & \textbf{Discrete parameters}       & \multicolumn{2}{c|}{\textbf{Value}} & \textbf{Unit} \\ 
		\hline
		$p_{30}$  & Slots per pole per phase         & \multicolumn{2}{c|}{[2-4]}          & -         \\ \hline
		$p_{31}$  & Pole pairs (p)                   & \multicolumn{2}{c|}{[3-5]}          & -         \\ \hline
        $p_{32}$  & Stator winding connection        & \multicolumn{2}{c|}{star/delta}          & -     \\ \hline
        $p_{33}$  & Winding scheme                   & \multicolumn{2}{c|}{short pitch/full pitch}          & -   \\ \hline
	\end{tabularx}%
\end{table}

We selected $T_{\textrm{PMSM}}:= 51532$ valid designs from the initial population. The number of samples for this dataset is close to the number of the previous ASM dataset ($\le3\%$); otherwise, the network can be biased towards one machine technology during training. The number of varying design parameters for PMSM is $d_{2}:=33$. The PMSM dataset also incorporates variability for topological and electrical parameters. A few crucial parameters listed in \autoref{tab:input_psm}  can be observed in \autoref{fig:TOP_PSM}, and representative samples with the varying pole pairs are illustrated in \autoref{fig:dvs}. The distribution of the listed parameters and KPIs is illustrated in \autoref{fig:PDS2} and \autoref{fig:KPIDS2}, respectively.

Our goal is to perform concurrent multi-objective optimization for both technologies. Hence, both datasets were generated under the assumption of identical KPIs and constant system parameters. The KPIs and constant system parameters information are given in \autoref{tab:kpis_info} and \autoref{tab:bcond}, respectively. Furthermore, during data generation, it is assumed that the cost of standard materials (such as aluminium and copper) is the same for ASM and PMSM. The significant difference in cost comes when considering magnets for PMSM.

\subsection{Training procedure}

 We employ a VAE based workflow, it maps the input data to a low-dimensional latent space through a probabilistic encoder and then reconstructs it through a decoder\cite{MAL-056}. As explained in \cite{9745548}, we first create a combined design space by concatenating the design vectors of all the given machines.
It creates $d$-dimensional design vector with $d= 1+d_1+\ldots+d_{M}$, where $d_1,\ldots,d_{M}$ represent input dimension of each different machine technology $t=1,\ldots,M$.  

After concatenation, we define any $i$th sample in the combined dataset as a $d$ dimensional vector if it is from the technology $t$ as
$$
\mathbf{p}^{(i)}=[t,\mathbf{0},\ldots,\mathbf{0},\mathbf{p}_{t}^{(i)},\mathbf{0},\ldots ,\mathbf{0}]
$$
and KPIs vector for the design as $\mathbf{k}^{(i)}=\mathbf{\kpi}_t(\textbf{p}_{t}^{(i)}).$

The total combined input dataset with all the machine types can be mathematically described by
\begin{align}
\mathcal{D} :=\Big\{ \mathbf{p}^{(i)} \;\Big|\; \text{for } i=1,\ldots,T_{\textrm{tot}}\, \Big\}
\end{align}
 The assumption is that $l$ dimensional unseen variables $\mathbf{z}$ from a latent distribution can describe all the $d$ dimensional input samples from the dataset $\mathcal{D}$. It is also assumed that the input parameters of each machine type are independent of each other. Therefore, as described in \cite{9745548}, to reconstruct parameters with high accuracy, the latent dimension $l$ should be at least higher than the maximum input dimension of all the machine types, i.e., $l \geq \max_M (d_M)$ and also $l \leq d $.

The encoder network computes the conditional distribution $\mathbf{P}(\mathbf{z}|\mathbf{p}) $ with the presumption that $\mathbf{z}$ follows the standard normal distribution. It is written as
\begin{align}
        \label{eq:encoder}
        ( \bm{\upsilon},\bm{\sigma}) := \mathbf{E}_{\phi}(\mathbf{p})
\end{align}
Where outputs mean ($\bm{{\upsilon}}$) and diagonal component of the covariance matrix as a vector ($\sigma$) represent latent distribution parameters with the dimension $l$. $\phi$ are trainable encoder network ($\mathbf{E}_{\phi}$) parameters.

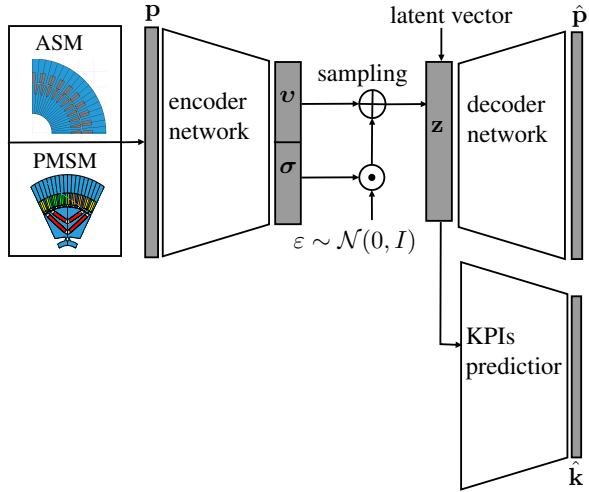
\begin{figure}[h!]
	\centering
	\input{fig/workflow_training_vae}
	\caption{VAE-based training workflow \cite{9745548}.}
	\label{fig:vae}
\end{figure}

To compute and backpropagate gradients  during training, as described in \cite{kingma2014auto}, the latent vector $\mathbf{z}$ is sampled using a reparametrization trick by 
\begin{align}
    \mathbf{z} = \bm{\upsilon} + \bm{\sigma} \odot   \bm{\varepsilon}
\end{align}
where $\bm{\varepsilon} \sim \mathcal{N}(0, \mathbf{I})$ is a noise vector, and $\odot$ is the component-wise dot product. The decoder network $\mathbf{D}_{\theta}$ takes latent vector $\mathbf{z}$ as input. It approximates the conditional distribution $\mathbf{P}(\mathbf{p}|\mathbf{z})$  i.e.,
\begin{align}
\hat{\mathbf{p}} := \mathbf{D}_{\theta}(\mathbf{z})
\end{align}
where $\theta$ are the trainable parameters of the decoder. Simultaneously, we train DNN with latent input ($\mathbf{z}$) to predict the KPIs. It is written as
\begin{align}
    \hat{\mathbf{k}} := \mathbf{K}_{\beta}(\mathbf{z})
\end{align}
where $\beta$ are trainable parameters of the DNN  and $\hat {\mathbf{k}}$ is vector of KPIs prediction.
The primary goal of training the VAE is to minimize the errors in prediction, parameter reconstruction, and encoding process by optimizing the network parameters $\theta,\phi, \beta$ simultaneously. 
One important step of before training is defining the loss function. The choice of the training loss function depends on the specific task at hand. The MSE is commonly used as a loss function for regression tasks \cite[Chapter 5]{Goodfellow-et-al-2016}, although other options, such as mean absolute error (MAE), exist. Through experiments, we determined that the MSE provides better prediction accuracy for our datasets. The MSE is a practical selection for parameter reconstruction since the input data is scalar. The total training loss comprises three components: the first two terms are squared error for parameter reconstruction and KPIs prediction, and the third term is Kullback-Leibler (KL) divergence for regularization in the latent space. Total VAE training loss is specified in terms of network parameters, input vector $\mathbf{p}^{(i)}$, and actual KPIs $\mathbf{k}^{(i)}$ by
\resizebox{\linewidth}{!}{%
\begin{minipage}{\linewidth}
\begin{align}
\mathbf{L}(\theta,\phi,\beta;(\mathbf{p}^{(i)},\mathbf{k}^{(i)})) 
    &=  \left \|\mathbf{p}^{(i)} - \hat{\mathbf{p}}^{(i)} \right \|^{2} 
    + \left \|\mathbf{k}^{(i)} -  \hat{\mathbf{k}}^{(i)} \right \|^{2} \nonumber
\\
    &+ \mathbf{D}_{\mathrm{KL}}\Big(\mathbf{P}(\mathbf{z}^{(i)}|\mathbf{p}^{(i)},\theta) \;||\; \mathbf{z} \sim \mathcal{N}(0,\mathbf{I})\Big)\nonumber
\end{align}
\end{minipage}
}
The KL divergence $\mathbf{D}_{\textrm{KL}}$ minimizes the difference between encoder distribution and prior distribution over latent variables. It works as a regularizer term in the loss function to provide continuity and completeness in the latent space. It means the samples nearby in the latent space remain similar when decoded while preserving meaningful representation \cite{amini2021mit}.
Here, for independent training of the DNN in a supervised manner\cite{9333549} for each machine technology,  we input $\mathbf{p}_t$ as the input vector instead of latent vector $\mathbf{z}$. It is written as
\begin{align}
    \hat{\mathbf{k}}_t := \mathbf{K}_{\gamma}(\mathbf{p}_t)
\end{align}
Here, $\gamma$ are network parameters, $\hat {\mathbf{k}}_t$ is the predicted KPIs vector, and $\mathbf{p}_t$ is an input vector of the individual machine technology with dimension $d_{t}$. The loss function during the network training is kept the same (MSE) as of the VAE for KPIs prediction. The only obvious change with the network structure is the input layer, compared to the used DNN during the VAE training. All the networks are trained using a standard back-propagation algorithm\cite{rumelhart1986learning}. \autoref{fig:vae} describes training workflow.
\subsection{MOO}
Every electrical machine design optimization comprises many design variables, constraints, and competing objectives (see \autoref{tab:kpis_info}). It leads to the generalized MOO problem formulation

\begin{align}
\label{eq:opt1}
\min_{\mathbf{p}}\quad & {k}_{a}(\mathbf{p}),  & {a}=1,\dots,n_{\mathrm{obj}}\\\label{eq:opt2}
\text{s.t.}\quad & {c}_{j}(\mathbf{p}) \leq 0, & j=1,\dots,n_{\mathrm{cons}}\\
&p_{i}^{\mathrm{L}} \leq p_{i} \leq p_{i}^{\mathrm{U}},\quad  \   & i=1,\dots,n_{\mathrm{param}}\label{eq:opt3}
\end{align}

where $\mathbf{p}$ represent input vector and ${p}_{i}^{\mathrm{L}}$ and ${p}_{i}^{\mathrm{U}}$ are parameter bounds, ${k}_{a}(\mathbf{p})$ characterise KPIs, ${c}_{j}(\mathbf{p})$ are constraints for design evaluation. Any commonly practiced multi-objective optimizer\cite{996017} can solve (\ref{eq:opt1}-\ref{eq:opt3}).

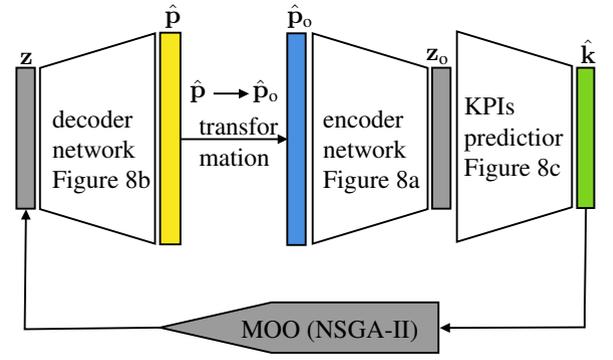
\begin{figure}[!h]
	\centering
	\input{fig/vae_optimization_workflow}
	\caption{Proposed VAE-based optimization workflow.}
	\label{fig:optimization_vae}
\end{figure}

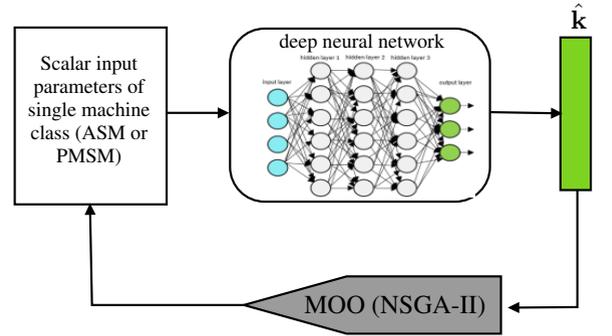
\begin{figure}[!h]
	\centering
	\input{fig/DNN_optimization_workflow}
	\caption{Individual DNN based optimization workflow.}
	\label{fig:optimization_dnn}
\end{figure}

In this study, we propose two MOO workflows: one for MTOO via continuous latent space using VAE (see \autoref{fig:optimization_vae}) and another (\autoref{fig:optimization_dnn}) for DNN-based classical workflow for individual machine technology.

In the VAE-based MTOO, randomly generated latent vector $\mathbf{z}$ is input to the optimization process. As shown in \autoref{fig:optimization_vae}, first, we give the latent vector as input $\mathbf{z}$ to the decoder($\mathbf{D}_{\theta}$), and the decoder predicts design parameters for the related technology. From the data pre-processing, we have prior knowledge about the positions of the actual design parameters in the concatenated form of the input vector. Hence, we keep those values unchanged and fill the remaining values with zero except for the first entry. The first entry indicates technology type, so we replace the predicted continuous value with a known integer value. We also replace predicted discrete parameters, such as pole pairs, slots per pole per phase, winding scheme, known zero values etc., with integer values based on prior knowledge. This makes the input vector ($\hat{\mathbf{p}}$) in the form ($\hat{\mathbf{p}}_{\textrm o}$) that the encoder ($\mathbf{E}_{\phi}$) expects. Then the encoder network creates a new latent vector $\mathbf{z}_{\textrm o}$, which is input to the KPIs predictor.
The standard DNN-based MOO workflow (\autoref{fig:optimization_dnn}) operates on one machine technology at a time, so we input a design vector with parameterization $\mathbf{p}_t\in\mathbb{P}_{t}\subset\mathbb{R}^{d_t}$. 

\section{Network architecture and Training specifications}\label{sec:NAT}

\subsection{Network architecture}

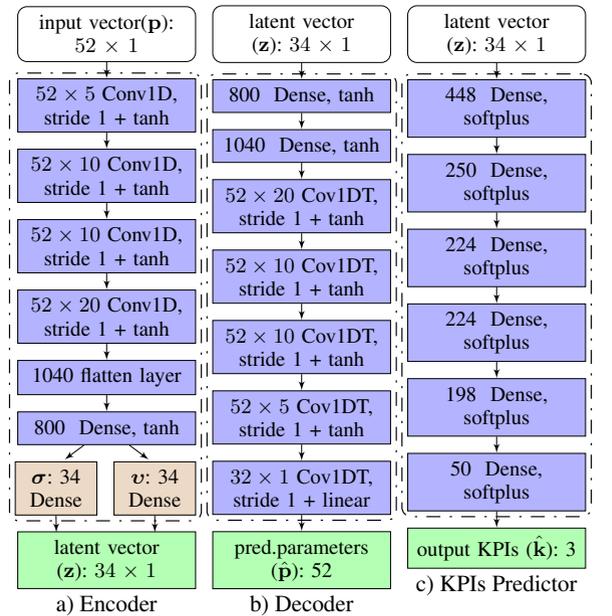
\begin{figure}[!h]
 	 \centering
 	 \input{fig/network_architecture}
     \caption{Network structure.}
     \label{fig:nat}
\end{figure}

 As illustrated in \autoref{fig:nat}, there are three networks: encoder ($\mathbf{E}_{\phi}$), decoder ($\mathbf{D}_{\theta}$) and the DNN ($\mathbf{K}_{\beta}$). The network structure and training hyperparameters are obtained randomly through trial and error by evaluating approximately twenty configurations starting with the base network configuration from \cite{9745548}. The details are as follows,

\begin{itemize}
     \item \textit{Encoder network}: The encoder($\mathbf{E}_{\phi}$) or inference network consists of four $1-d$ convolutional layers. These layers are significant for learning essential features from the combined design vector and determining whether the design is ASM or PMSM from the technology indicator (the first input parameter of the design vector). A flattened layer and a dense layer follow these convolutional layers. Three output layers follow the dense layer. Two output layers are the distribution parameters mean ($\bm{\upsilon}$) and variance ($\bm{\sigma}$) that sample the final output latent vector ($\mathbf{z}$). 
      \item \textit{Decoder network}: The decoder network ($\mathbf{D}_{\theta}$), also called the generative network, consists of two dense layers, four $1-d$ convolution transposed layers, and one output layer corresponding to the dimension of the integrated design space. It follows the reverse structure of the encoder except for the output layer. The output layer has a linear activation function. 
      \item \textit{DNN}: The DNN is also interchangeably known as KPIs predictor in this study. The DNN has one input, five dense layers, and one output layer. For individual DNNs, we use an identical network structure (except the input layer) and hyperparameters, where we only train the DNN for one machine type at a time.
\end{itemize}

\subsection{Training specifications}

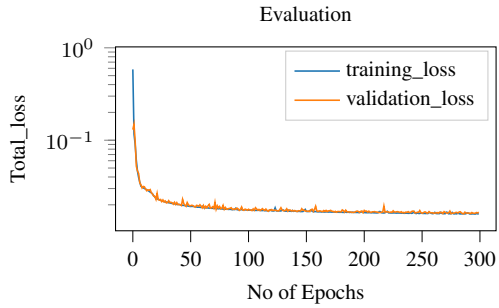
\begin{figure}
 	\centering
 	\input{fig/validation_loss}
    \caption{Training and validation loss curves.}
    \label{fig:val_loss}
\end{figure}

 \begin{table}
	\caption{Training hyperparameters detail}
	\label{tab:HPO_info}
	\newcolumntype{Y}{>{\centering\arraybackslash}X}
	\begin{tabularx}{\linewidth}{|Y|X|Y|Y|}
		\hline
        \textbf{Parameters}       & \multicolumn{1}{c|}{\textbf{Value}}\\ 
		\hline
	    Learning rate                       & \multicolumn{1}{c|}{$10^{-4}$-$10^{-5}$}\\ \hline
	    Total number of epochs              & \multicolumn{1}{c|}{300} \\ \hline
		Validation patience                 & \multicolumn{1}{c|}{20}\\ \hline
		Optimizer                           & \multicolumn{1}{c|}{Adam\cite{kingma2015adam}}\\ \hline
		Latent space dimension              & \multicolumn{1}{c|}{34 (as $d_{\textrm {PMSM}}:=33$)}\\ \hline
		Loss functions                      & \multicolumn{1}{c|}{KL-divergence and MSE }\\ \hline
		Batch size                          & \multicolumn{1}{c|}{50}\\ \hline
	\end{tabularx}%
\end{table}

 The total number of samples in a combined dataset is $T_{\mathrm{tot}}:= T_{\mathrm{ASM}}+T_{\mathrm{PMSM}}:=101919$. We split a total number of samples ($T_{\mathrm{tot}}$) into three sets: training ($\sim 80 \%$ of $T_{\mathrm{tot}}$) , validation ($\sim 10 \%$), and testing ($\sim 10 \%$). \autoref{fig:val_loss} displays training and validation curves. \autoref{tab:HPO_info} gives the details of the training hyperparameters. The network training was carried out on NVIDIA Quadro M2000M GPU. It took $\sim 1.5$ hours to complete the VAE training for the multi-technology scenario, whereas separate DNNs takes around  $\sim 15$ minutes with the for single machine technology. The magneto-static FE simulation takes $2-4$\,h/sample for PMSM, and analytical calculation takes around $5-7$ minutes/sample for ASM on a single-core CPU.

\section{Numerical results}\label{sec:NR}

\begin{figure}[!h]
    \centering 
    \includegraphics[width = \linewidth]{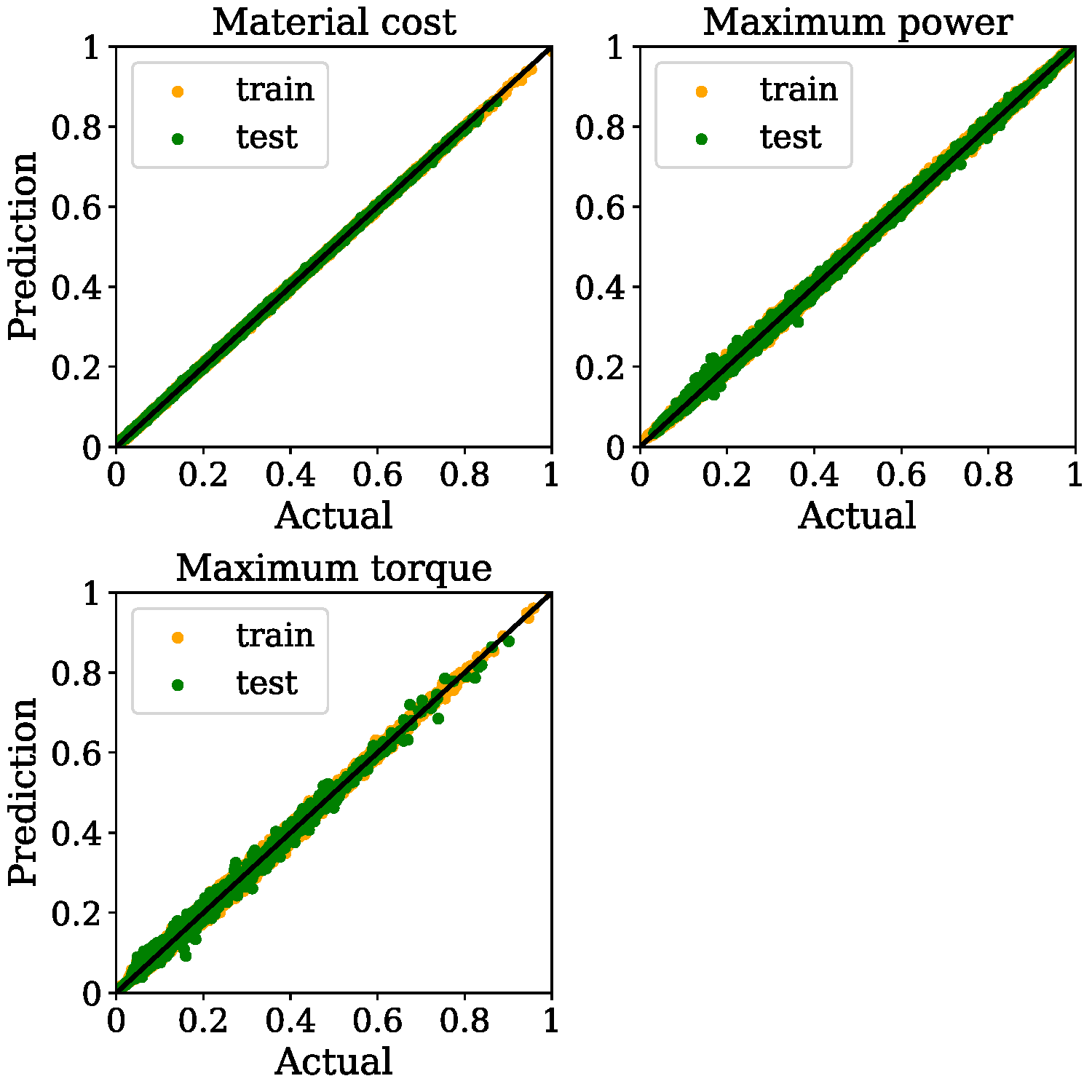}
    \caption{Predictions of the KPIs over test samples.}
    \label{fig:kpi_pred}
\end{figure}

\begin{figure*}
    \centering                  
    \includegraphics[width=0.8\linewidth]{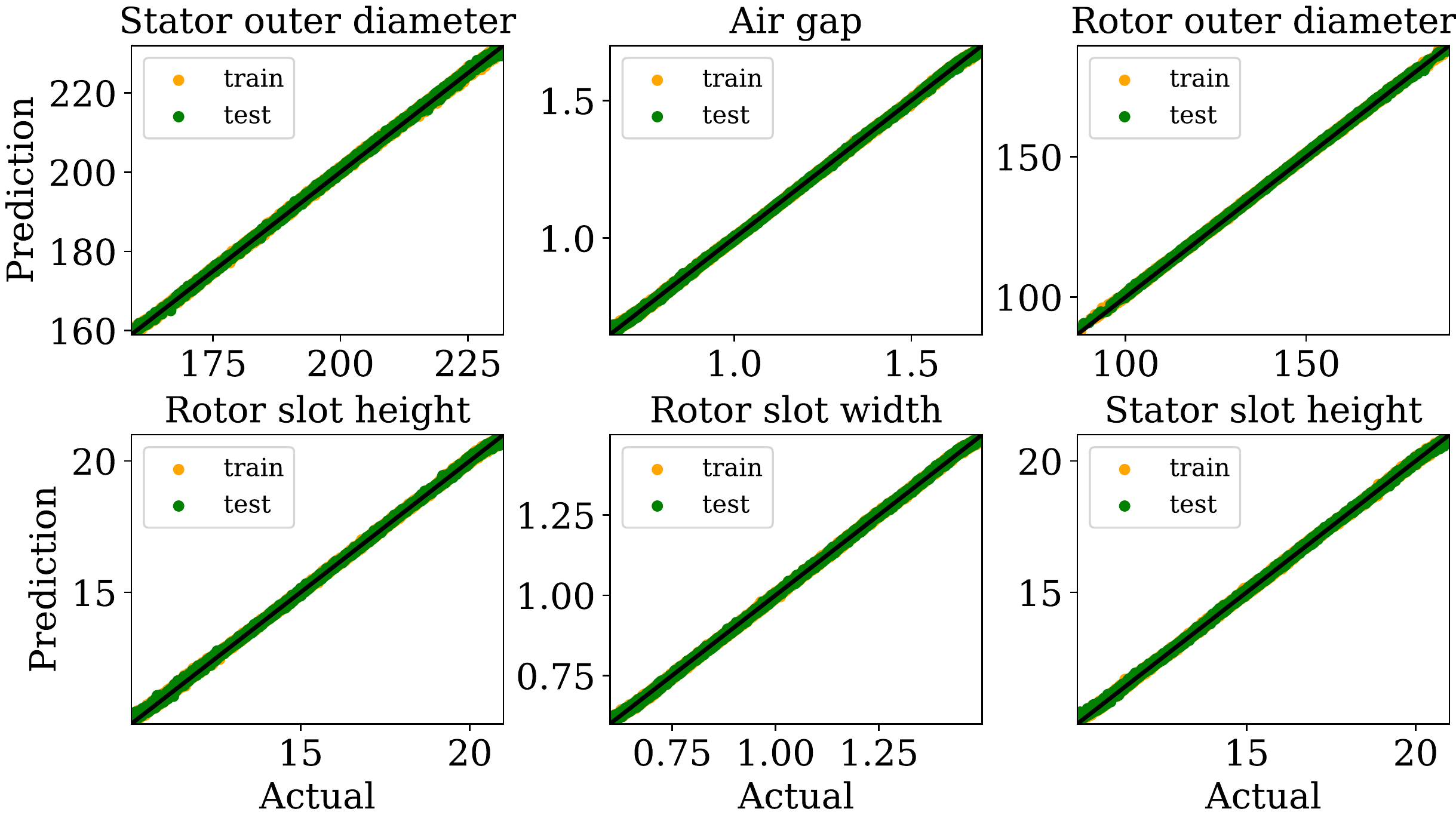}
    \caption{ASM parameters prediction plot over test samples.}
    \label{fig:param_plot_asm}
\end{figure*}
 \begin{figure*}
    \centering                  
    \includegraphics[width=0.8\linewidth]{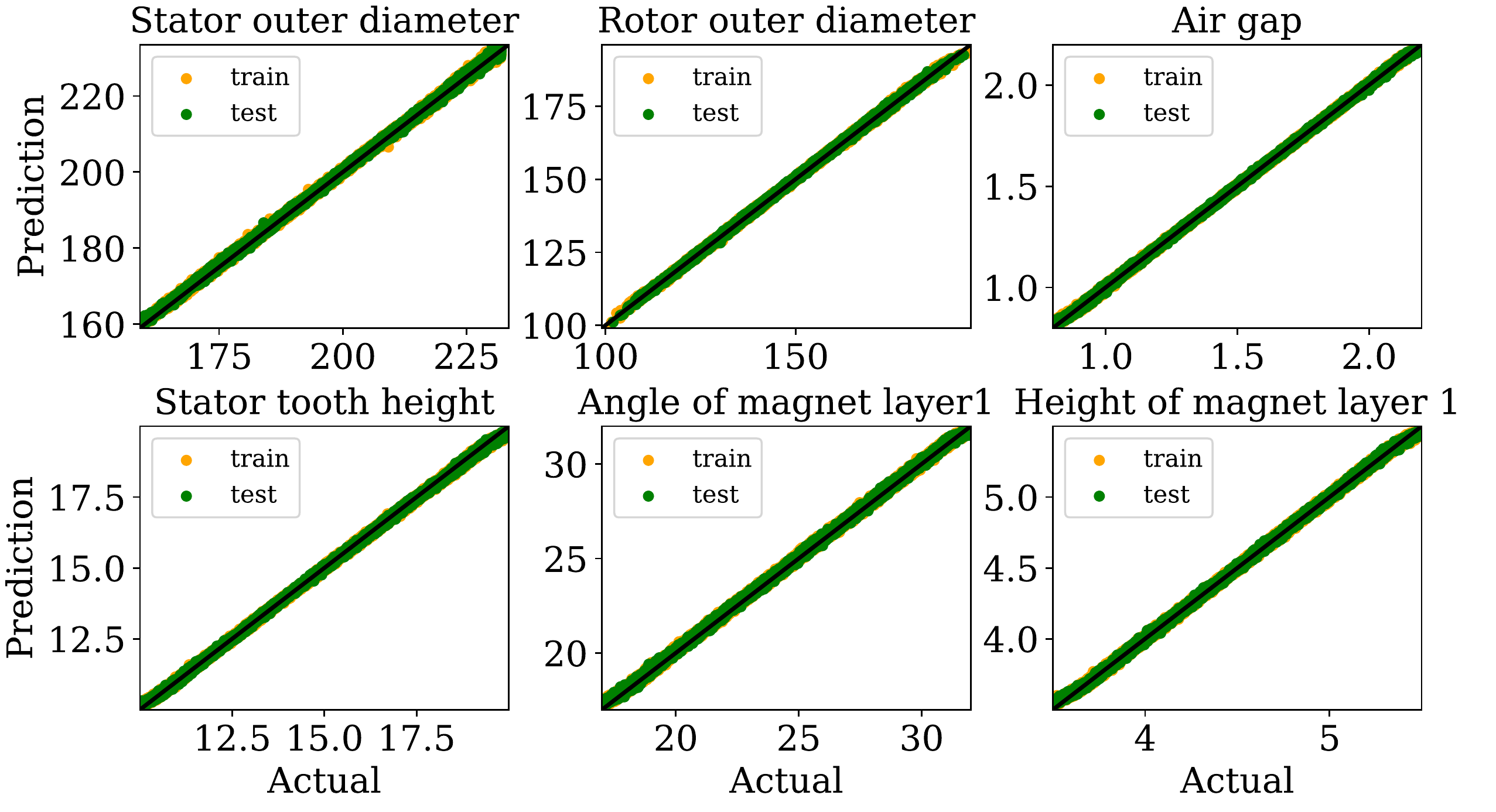}
    \caption{PMSM parameters prediction plot over test samples.}
    \label{fig:param_plot_psm}
\end{figure*}

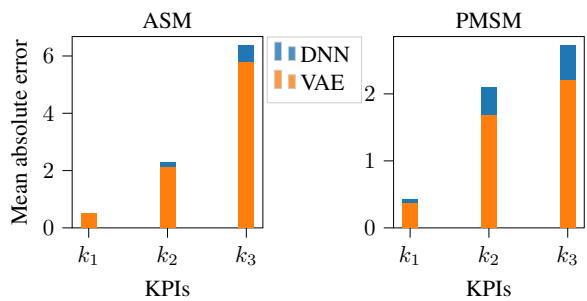
\begin{figure}
        \centering
        \input{fig/vae_mlp_cmp}
        \caption{Comparison VAE vs DNN.}
        \label{fig:cmp_dnn_vae}
\end{figure}%

\begin{table}[h!]
\caption{KPIs evaluation over test samples}
\label{tab:KPIs}
\setlength{\tabcolsep}{10pt}
\renewcommand{\arraystretch}{1.2}
\footnotesize
\resizebox{\linewidth}{!}{%
\begin{tabular}{|c|c|c|c|c|}
\hline
\multirow{2}{*}{}   & \multicolumn{4}{c|}{\textbf{Prediction accuracy}} \\ \cline{2-5} 
                    & \textbf{MAE} & \textbf{RMSE} & \textbf{PCC} & \textbf{MRE(\%)} \\ \hline
$\kpi_{ 1}$         & 0.43         & 0.53          & 1            & 0.71             \\ \hline
$\kpi_{ 2}$         & 1.90         & 2.54          & 1            & 1.31             \\ \hline
$\kpi_{ 3}$         & 3.96         & 6.52          & 0.99         & 1.76             \\ \hline
\end{tabular}%
}
\end{table}

\begin{table}[]
\caption{ASM parameters evaluation over test samples}
\label{tab:asm_p}
\setlength{\tabcolsep}{4pt}
\renewcommand{\arraystretch}{1.2}
\normalsize
\resizebox{\linewidth}{!}{%
\begin{tabular}{|c|c|c|c|c|}
\hline
  \multirow{2}{*}{\textbf{Parameters}} &
  \multicolumn{4}{c|}{\textbf{Reconstruction accuracy}} \\ \cline{2-5} 
                                   & \textbf{MAE} & \textbf{RMSE}  & \textbf{PCC} & \textbf{MRE(\%)} \\ \hline
 Stator outer diameter     & 0.39  & 0.483 & 0.99 & 0.19              \\ \hline
 Air gap                   & 0.002 & 0.005 & 1    & 0.37              \\ \hline
 Rotor outer diameter      & 0.55  & 0.636 & 0.99 & 0.41              \\ \hline
 Rotor slot height         & 0.05  & 0.069 & 0.99 & 0.37              \\ \hline
 Rotor slot width          & 0.004 & 0.005 & 1    & 0.4               \\ \hline
 Stator slot height        & 0.05  & 0.004 & 1    & 0.32              \\ \hline
\end{tabular}%
}
\end{table}

\begin{table}[]
\caption{PMSM parameters evaluation over test samples}
\label{tab:psm_p}
\setlength{\tabcolsep}{4pt}
\renewcommand{\arraystretch}{1.2}
\normalsize
\resizebox{\linewidth}{!}{%
\begin{tabular}{|c|c|c|c|c|}
\hline
  \multirow{2}{*}{\textbf{Parameters}} &
  \multicolumn{4}{c|}{\textbf{Reconstruction accuracy}} \\ \cline{2-5} 
                                   & \textbf{MAE} & \textbf{RMSE} & \textbf{PCC} & \textbf{MRE(\%)} \\ \hline
 Stator outer diameter             & 0.51         & 0.63          & 0.99         & 0.26             \\ \hline
 Rotor outer diameter              & 0.39         & 0.5           & 0.99         & 0.27             \\ \hline
 Air gap                           & 0.006        & 0.007         & 1            & 0.41             \\ \hline
 Stator tooth height               & 0.054        & 0.072         & 0.99         & 0.37             \\ \hline
  Angle magnet layer 1             & 0.11         & 0.14          & 0.99         & 0.48             \\ \hline
 Height of magnet layer 1          & 0.012        & 0.015         & 1            & 0.27             \\ \hline
\end{tabular}%
}
\end{table}

\begin{table}
\caption{MOO settings information}
\label{tab:opt_hp_info}
\setlength{\tabcolsep}{4pt}
\renewcommand{\arraystretch}{1.2}
\normalsize
\newcolumntype{Y}{>{\centering\arraybackslash}X}
\begin{tabularx}{\linewidth}{|Y|X|Y|Y|}
	\hline
    \textbf{Settings}       & \multicolumn{1}{c|}{\textbf{Value}}\\ 
	\hline
    Sampling approach                       & \multicolumn{1}{c|}{random initialization}\\ \hline
    Population per generation               & \multicolumn{1}{c|}{1000} \\ \hline
	Stopping criteria                       & \multicolumn{1}{c|}{100 generations}\\ \hline
	Number of objectives                    & \multicolumn{1}{c|}{2}\\ \hline
	Crossover, mutation probability                 & \multicolumn{1}{c|}{0.9}\\ \hline
\end{tabularx}%
\end{table}

\begin{figure*}
        \centering
        \input{fig/Pareto_design_visualization}
        \caption{Pareto designs.}
        \label{fig:Pareto_vae_dnn}
\end{figure*}
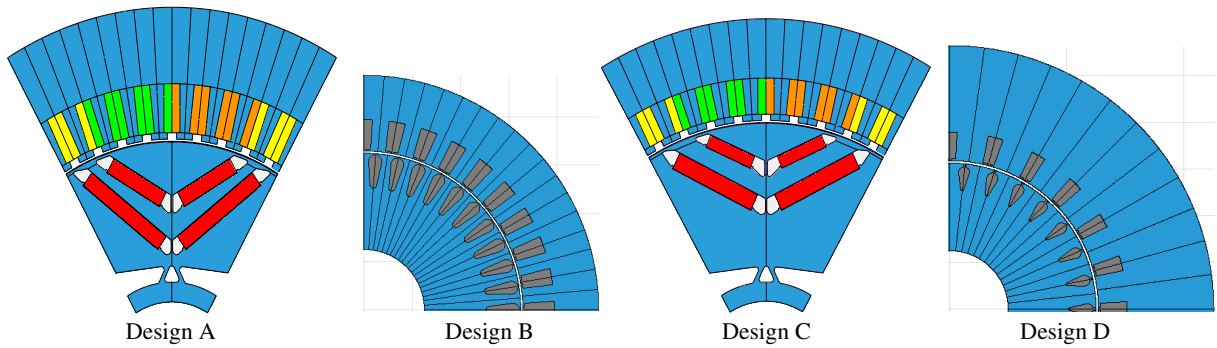%

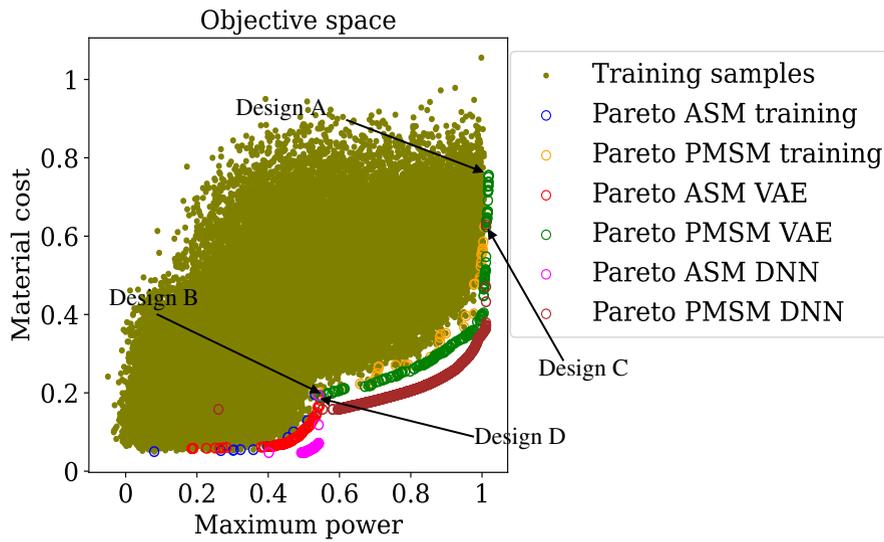
\begin{figure*}
 	\centering
 	\input{fig/Pareto_front_all_combined}
    \caption{Pareto-fronts for Maximum power and Material cost, where model training samples are in olive, the Pareto-front of ASM training samples is in blue, and the Pareto-front of PMSM training samples is in orange.  Pareto-fronts for the VAE-based approach are displayed in red (PMSM), and green (ASM), and Pareto-fronts for the DNN-based direct approach are shown in brown (PMSM) and magenta (ASM).}
    \label{fig:moo}
\end{figure*}
 
\begin{table}[]
\caption{Design evaluation from VAE Pareto front}
\label{tab:pareto_pred_vae}
\setlength{\tabcolsep}{4pt}
\renewcommand{\arraystretch}{1.2}
\normalsize
\resizebox{\linewidth}{!}{%
\begin{tabular}{|c|c|c|c|c|c|c|}
\hline
\multirow{2}{*}{\textbf{KPIs}} & \multicolumn{3}{c|}{\textbf{Design A (PMSM) }}                & \multicolumn{3}{c|}{\textbf{Design B (ASM)}}                \\ \cline{2-7} 
                           & \textbf{FE simulation} & \textbf{Prediction} & \textbf{MRE(\%)} & \textbf{FE simulation} & \textbf{Prediction} & \textbf{MRE(\%)} \\ \hline
$\kpi_{ 1}$                             & 153.53                  & 153.49               & 0.026            & 45.47                   & 46.27                & 1.75             \\ \hline
$\kpi_{ 2}$                             & 402.98                  & 406.70               & 0.92             & 241.7                   & 237.82               & 1.60             \\ \hline
$\kpi_{ 3}$                             & 294.61                  & 286.63               & 2.70             & 195.46                  & 187.52               & 4.06            \\ \hline
\end{tabular}%
}
\end{table}

\begin{table}[]
\caption{Design evaluation from DNNs Pareto front}
\label{tab:pareto_pred_dnn}
\setlength{\tabcolsep}{4pt}
\renewcommand{\arraystretch}{1.2}
\normalsize
\resizebox{\linewidth}{!}{%
\begin{tabular}{|c|c|c|c|c|c|c|}
\hline
\multirow{2}{*}{\textbf{KPIs}} & \multicolumn{3}{c|}{\textbf{Design C (PMSM) }}                & \multicolumn{3}{c|}{\textbf{Design D (ASM)}}                \\ \cline{2-7} 
                               & \textbf{FE simulation} & \textbf{Prediction} & \textbf{MRE(\%)} & \textbf{FE simulation} & \textbf{Prediction} & \textbf{MRE(\%)} \\ \hline
$\kpi_{ 1}$                             & 130.81                  & 129.51              & 0.99             & 46.68                 & 46.73                & 0.53             \\ \hline
$\kpi_{ 2}$                             & 401.8                   & 404                & 0.54              & 227.48                & 235                  & 3.3              \\ \hline
$\kpi_{ 3}$                             & 313.05                 & 316.27              & 1.02              & 158                   & 200.36               & 26.81              \\ \hline
\end{tabular}%
}
\end{table}

In this study, our primary focus is on the concurrent multi-technology scenario; therefore, we will explain the evaluation of the trained VAE in more detail. After training, we test trained models on the test dataset. \autoref{tab:KPIs} gives evaluation details of test samples for all three global KPIs. We use unitless mean relative error (MRE), the root mean squared error (RMSE), MAE, and Pearson correlation coefficient (PCC) to measure the correlation between input and target variables. \autoref{fig:kpi_pred} illustrates prediction plot of all these three KPIs over test samples. We can see that the maximum torque KPI ($k_3$) has a higher MAE ($3.96$ Nm). We present a numerical analysis of parameter reconstruction of six parameters of each machine type in \autoref{tab:asm_p} and \autoref{tab:psm_p}. It is observed that the parameter reconstruction is obtained with high precision. \autoref{fig:param_plot_asm} and \autoref{fig:param_plot_psm} display prediction plots. The other parameters, which are not illustrated here, also have higher reconstruction accuracy. 

\autoref{fig:cmp_dnn_vae} displays a numerical comparison of the KPIs prediction performance of VAE and DNN with MAE over the same test samples. For the numerical analysis, the DNN is trained with a twin network configuration and hyperparameters as used for the DNN for the latent input. The DNNs are directly trained on the input parameters of each machine type, using a supervised learning approach. The VAE has a slightly better prediction accuracy than the trained DNN for the single machine types. This is likely due to more training samples in the combined dataset and more accurate functional mapping between latent input and output global KPIs. 

The trained models (encoder, decoder and DNN) are used for MTOO for $T=2$ technologies. We propose an improved optimization workflow \autoref{fig:optimization_vae}. The proposed workflow improves synchronization between the decoder and the KPIs predictor in the optimization loop. It also handles many discrete input parameters effectively. 

We perform MOO for two competing global KPIs: material cost and maximum power. We use the genetic algorithm NSGA-\rom{2}, which can handle many design variables \cite{996017}. We set the MOO hyperparameters by experience; see \autoref{tab:opt_hp_info} for standard hyperparameters. The optimization process includes input parameter bounds as a constraint to reduce invalid design generation. The optimization requires roughly  $\sim 2.5$ hours. We also run individual MOO for each machine type with the separately trained DNN model. Identical hyperparameter settings as of the VAE-based optimization are used. Each machine optimization takes around $\sim 40-50$ minutes. 
 \autoref{fig:moo} figure depicts different Pareto-fronts for the training samples, the VAE approach, and the separate DNN approach. All Pareto-fronts from DNNs and VAE show power or cost-effective designs. It can be seen that Pareto-fronts based on VAE and DNNs predictions contain designs that are not present in the training data. We illustrate two designs from each Pareto front. Design A (PMSM) and Design B (ASM) from the VAE Pareto front. Similarly, Design C (PMSM) and Design D (ASM) from the individually trained DNN models. We recalculated all these four designs with their conventional approach. \autoref{tab:pareto_pred_vae} shows the evaluation for all three KPIs with MRE. We can see that Design B (ASM) has a high ($4.06\%$) MRE for maximum torque KPI. Likewise, \autoref{tab:pareto_pred_dnn} evaluates Design C (PMSM) and  D (ASM). For Design D, MRE is very high, $26.81 \%$, possibly due to a poor approximation of the meta-model for the maximum torque for that design.
It is seen from \autoref{fig:moo} that the DNN-based separate approach shows a more efficiently obtained Pareto front. We checked twenty designs from each Pareto front of both approaches. For the direct DNN-based approach, most designs from the more efficiently obtained Pareto region were found geometrically invalid. We observe that, even if we apply input parameter bounds as constraints to lower invalid design generation, we get much higher invalid designs ($\sim 60\%$) compared to VAE-based concurrent optimization. If the design is valid, then the prediction has a high deviation after recalculating with the conventional approach. Design D from \autoref{fig:moo} is the example where we observe that recalculation produces a higher deviation in the prediction of torque KPI (MRE is $26.81\%$; see \autoref{tab:pareto_pred_dnn}). This can be improved by adding geometry checks during  optimization. However, that is beyond the scope of work.

\section{Conclusion and outlook}\label{sec:conclusion}

We present the application of the VAE-based approach for optimizing two different machine types (ASM and PMSM) simultaneously over a common set of KPIs, i.e., material cost and maximum power. The numerical results demonstrate high prediction accuracy for parameter reconstruction and KPIs in a complex design space. This enables the optimization of several electrical machine technologies with a single meta-model training. The quantitative analysis for the DNN-based direct approach for optimizing each machine type is also demonstrated. The MOO results show that direct DNN based approach has a more efficiently obtained Pareto region, but the VAE outputs more valid meaningful designs than independent optimization with DNN. However, the DNN-based optimization takes less computational effort than the VAE-based approach for fewer machines (in this study two). We expect a linear increase in the computational time during optimization for multiple machine types when the DNN-based models are trained separately. On the contrary, only a little increase in the computational time is expected for the VAE-based approach. Future work may include the application of trained meta-models in other query scenarios, such as sensitivity analysis and uncertainty quantification. The more challenging situation, e.g. optimization for more than two machine technologies, can also be considered.

\printbibliography

\begin{IEEEbiography}[{\includegraphics[width=1in,height=1.25in,clip,keepaspectratio]{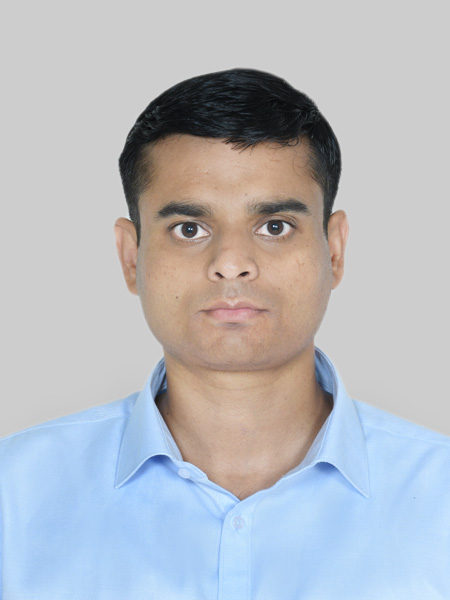}}]{Vivek Parekh} received his bachelor degree in electronics and communication from Gujarat technological university in 2012. He worked as an Assistant manager in Reliance Jio Infocomm Ltd. from 2013 to 2015 in operation and maintenance. He has obtained his M.Sc degree in information technology from the University of Stuttgart in 2019. Currently, he is pursuing a Ph.D. at TU Darmstadt in the Institut f\"{u}r Teilchenbeschleunigung und Elektromagnetische Felder under the fellowship of Robert Bosch GmbH. His area of interest comprises machine learning, deep learning, reinforcement learning, optimization of electrical machine design, and the industrial simulation process.
\end{IEEEbiography}
\begin{IEEEbiography}[{\includegraphics[width=1in,height=1.25in,clip,keepaspectratio]{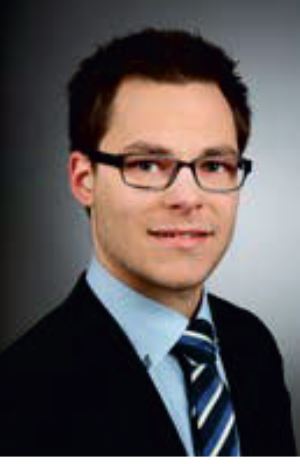}}]{Dominik Flore} received the bachelor in maschinenbau with specialization in mechatronik and master in maschinenbau with specialization in Product development from the University of Paderborn in 2012 and 2013, respectively. He obtained a Ph.D. degree with the topic "Experimentelle Untersuchung und Modellierung des Sch\"{a}digungsverhaltensfaserverst\"{a}rkter Kunststoffe" from ETH Z\"{u}rich, in 2016. Currently, he is working as a development engineer for the reliability of the electric machine at Robert Bosch GmbH, Stuttgart.
Present research interests involve product development in the field of electrical machine with the application of machine learning, optimization of the industrial simulation process.
\end{IEEEbiography}
\begin{IEEEbiography}[{\includegraphics[width=1in,height=1.25in,clip,keepaspectratio]{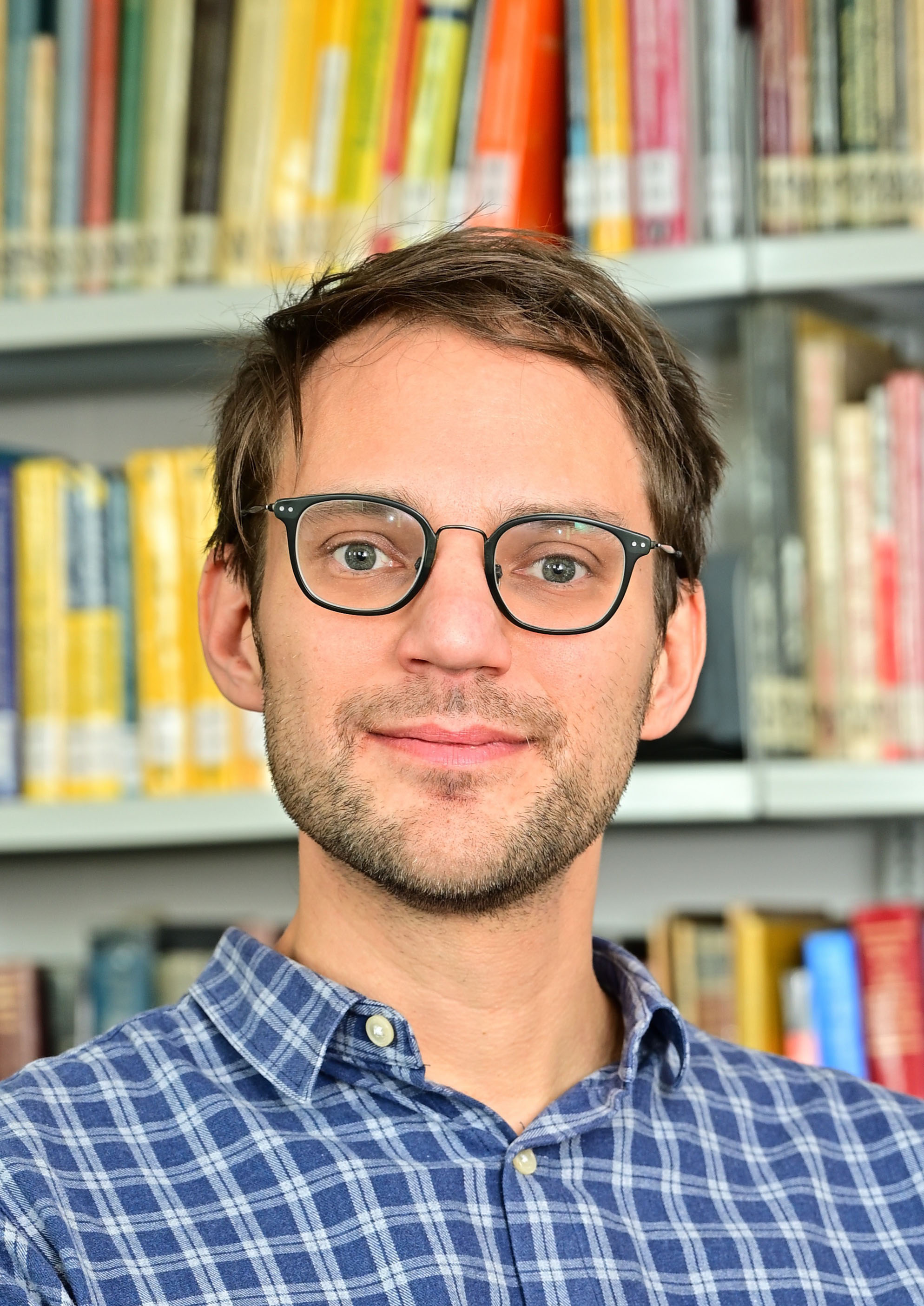}}]{Sebastian Sch\"{o}ps} received the M.Sc. degree in business mathematics and a joint doctoral degree from the Bergische Universität Wuppertal and the Katholieke Universiteit Leuven in mathematics and physics, respectively. He was appointed a Professor of computational electromagnetics with the Technische Universität Darmstadt within the interdisciplinary centre of computational engineering in 2012. Current research interests include coupled multiphysical problems, bridging computer aided design and simulation, parallel algorithms for high performance computing, digital twins, uncertainty quantification and machine learning.
\end{IEEEbiography}

\EOD
\end{document}

%% file: fig/data_visualization.tex
\tikzset{every picture/.style={line width=0.75pt}} 
\resizebox{\linewidth}{!}{
\begin{tikzpicture}[x=0.75pt,y=0.75pt,yscale=-1,xscale=1]

\draw (339.83,109.88) node  {\includegraphics[width=75.25pt,height=75.25pt]{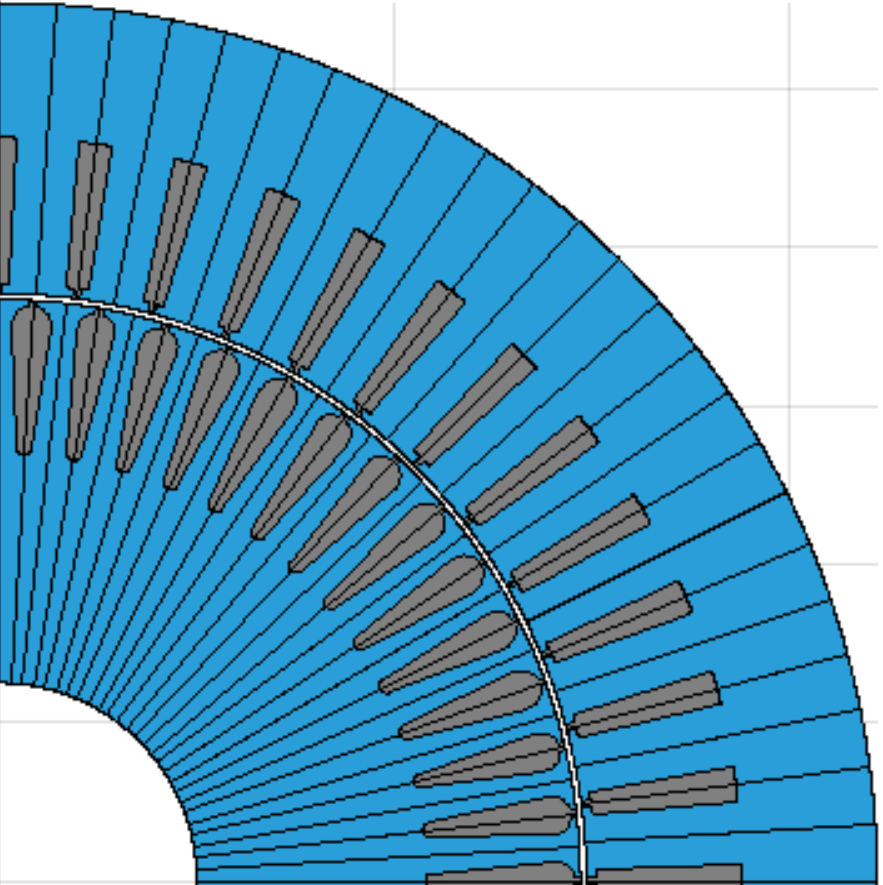}};
\draw (339.83,245) node  {\includegraphics[width=70.25pt,height=90pt]{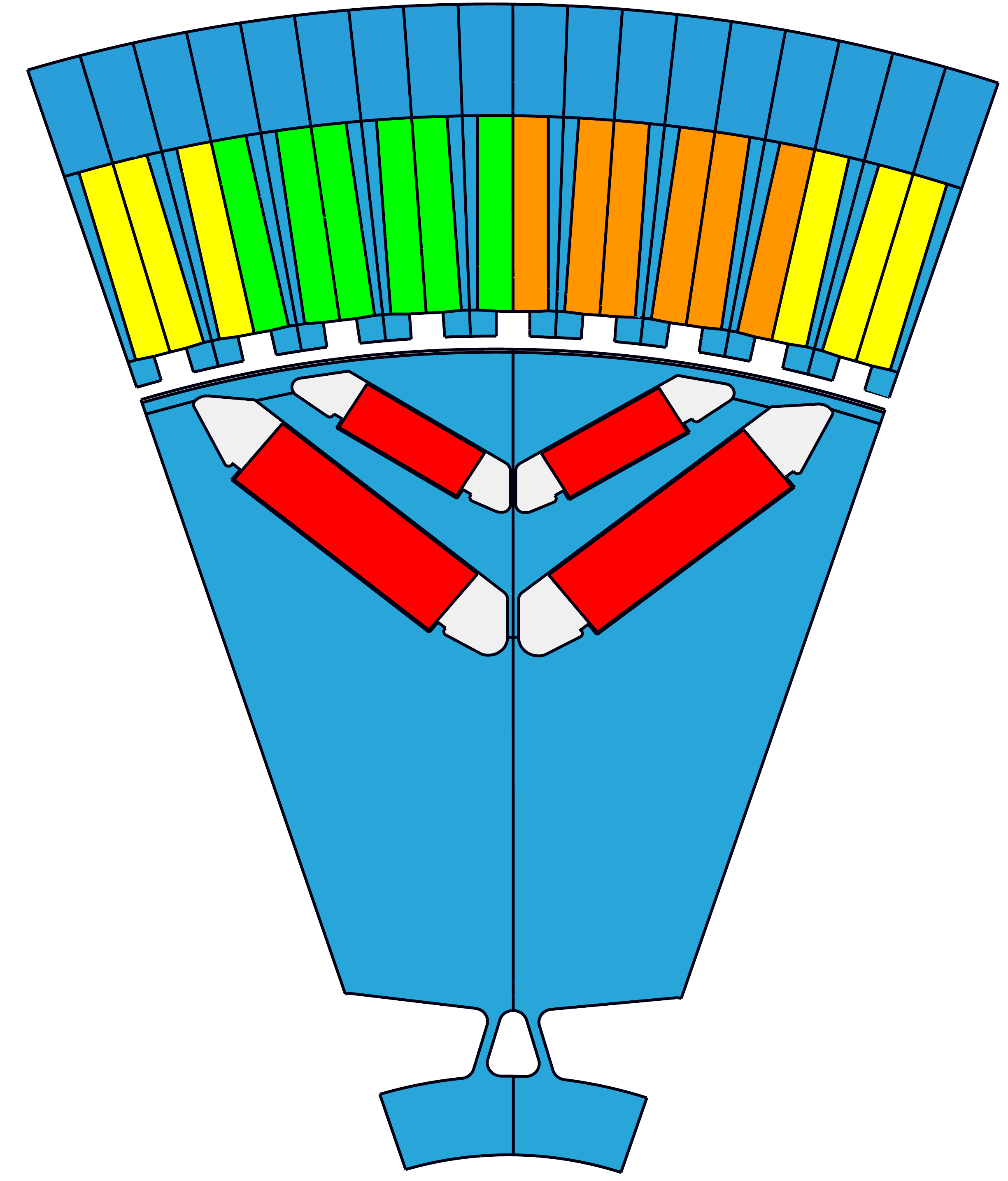}};
\draw (100.5,109.88) node  {\includegraphics[width=75.25pt,height=75.25pt]{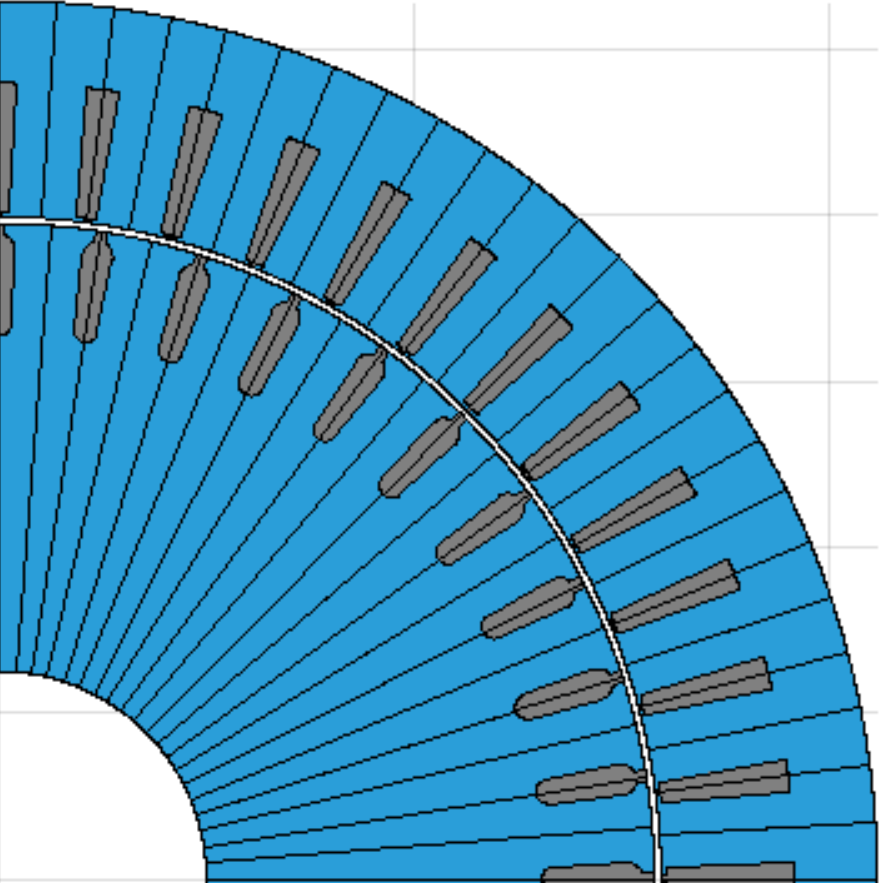}};
\draw (100.5,245) node  {\includegraphics[width=105.25pt,height=90pt]{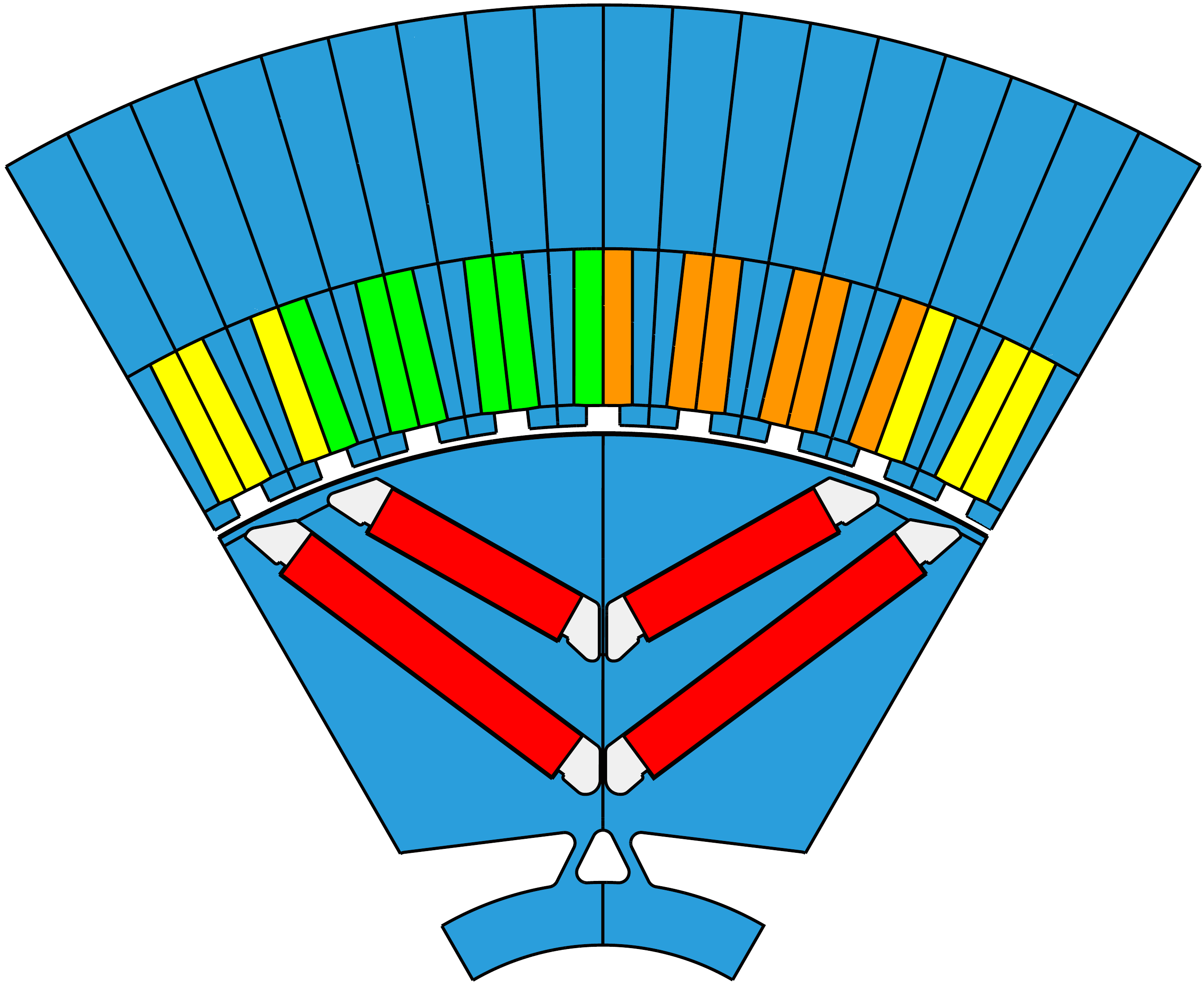}};
\draw (220.25,109.88) node  {\includegraphics[width=75.25pt,height=75.25pt]{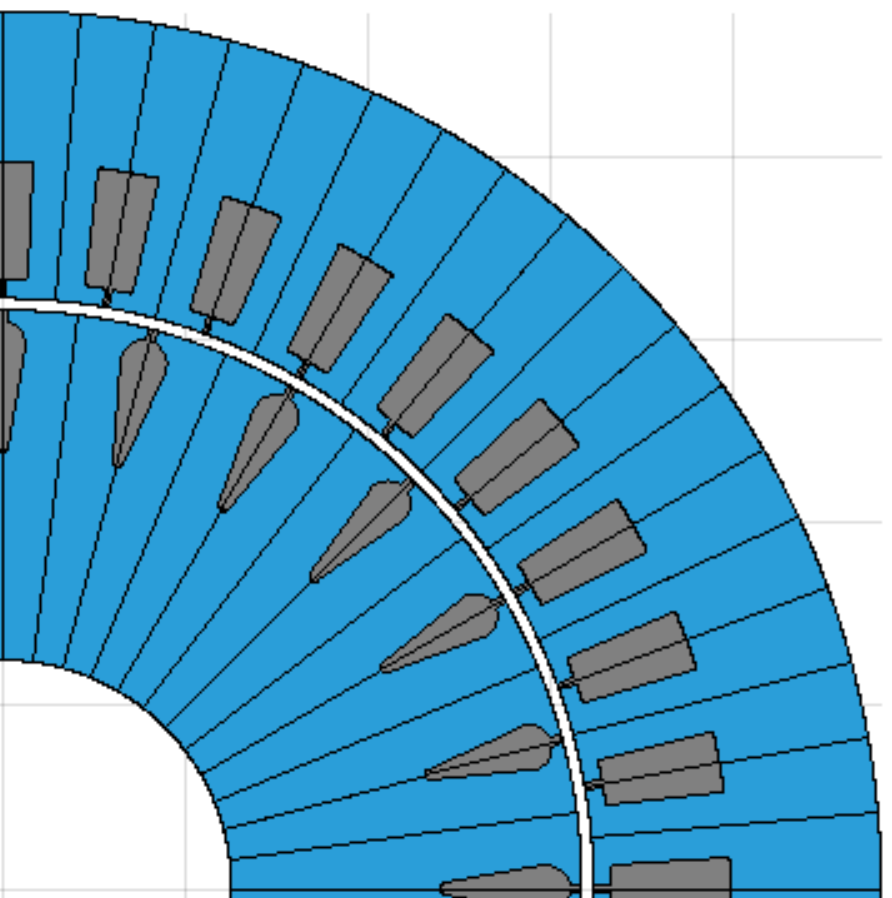}};
\draw (230,245) node  {\includegraphics[width=80pt,height=90pt]{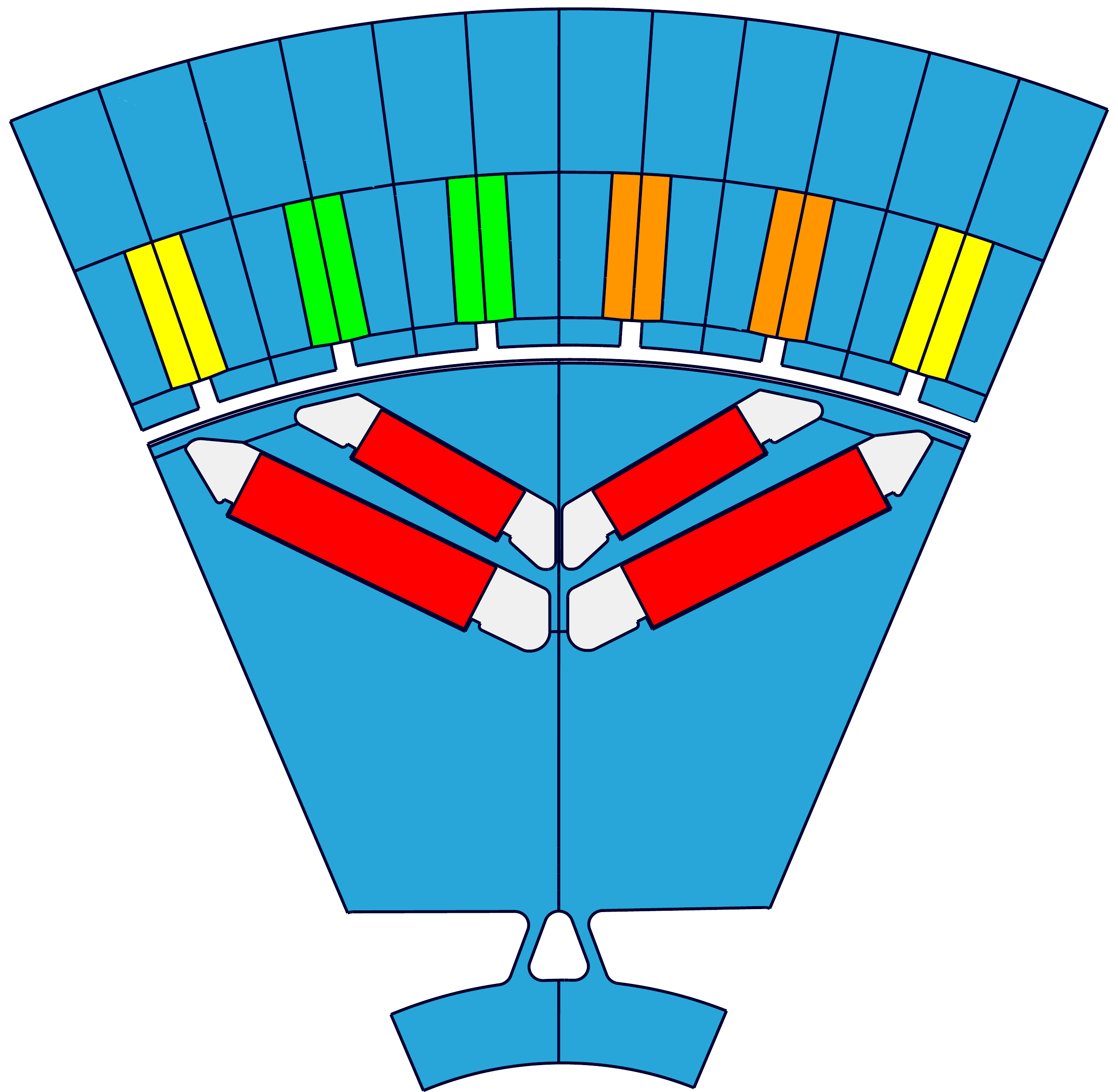}};

\draw (100,172.46) node  [font=\small] [align=left] {\begin{minipage}[lt]{25pt}\setlength\topsep{0pt}
a) p:2
\end{minipage}};
\draw (222.5,172.46) node  [font=\small] [align=left] {\begin{minipage}[lt]{25pt}\setlength\topsep{0pt}
b) p:3
\end{minipage}};
\draw (343,172.46) node  [font=\small] [align=left] {\begin{minipage}[lt]{25pt}\setlength\topsep{0pt}
c) p:4
\end{minipage}};
\draw (102,314) node  [font=\small] [align=left] {\begin{minipage}[lt]{25pt}\setlength\topsep{0pt}
d) p:3
\end{minipage}};
\draw (233,314) node  [font=\small] [align=left] {\begin{minipage}[lt]{25pt}\setlength\topsep{0pt}
e) p:4
\end{minipage}};
\draw (343,314) node  [font=\small] [align=left] {\begin{minipage}[lt]{25pt}\setlength\topsep{0pt}
f) p:5
\end{minipage}};

\end{tikzpicture}}

%% file: fig/asm_technology_sample.tex
\tikzset{every picture/.style={line width=0.75pt}} 
\resizebox{\linewidth}{!}{
\begin{tikzpicture}[x=0.75pt,y=0.75pt,yscale=-1,xscale=1]

\draw (237,145.5) node  {\includegraphics[width=189.5pt,height=189.38pt]{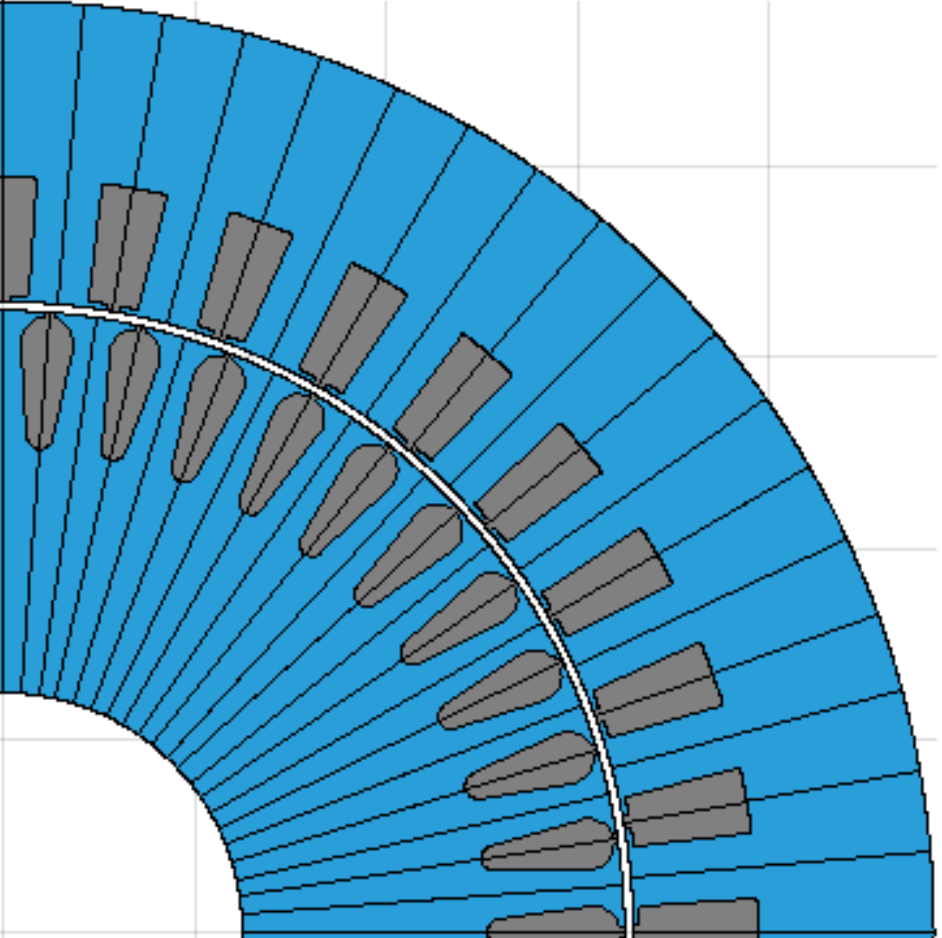}};
\draw    (107,23.21) -- (107,268.64) ;
\draw [shift={(106.90,271.64)}, rotate = 270] [fill={rgb, 255:red, 0; green, 0; blue, 0 }  ][line width=0.08]  [draw opacity=0] (5.36,-2.57) -- (0,0) -- (5.36,2.57) -- cycle    ;
\draw [shift={(106.90,20.21)}, rotate = 90] [fill={rgb, 255:red, 0; green, 0; blue, 0 }  ][line width=0.08]  [draw opacity=0] (5.36,-2.57) -- (0,0) -- (5.36,2.57) -- cycle    ;
\draw    (109.64,101.32) -- (109.61,103.79) ;
\draw [shift={(109.57,106.79)}, rotate = 270.7] [fill={rgb, 255:red, 0; green, 0; blue, 0 }  ][line width=0.08]  [draw opacity=0] (3.57,-1.72) -- (0,0) -- (3.57,1.72) -- cycle    ;
\draw [shift={(109.67,98.32)}, rotate = 90.7] [fill={rgb, 255:red, 0; green, 0; blue, 0 }  ][line width=0.08]  [draw opacity=0] (3.57,-1.72) -- (0,0) -- (3.57,1.72) -- cycle    ;
\draw    (224.64,149.47) -- (114.03,268.87) ;
\draw [shift={(112.67,270.33)}, rotate = 312.81] [fill={rgb, 255:red, 0; green, 0; blue, 0 }  ][line width=0.08]  [draw opacity=0] (7.2,-1.8) -- (0,0) -- (7.2,1.8) -- cycle    ;
\draw [shift={(226,148)}, rotate = 132.81] [fill={rgb, 255:red, 0; green, 0; blue, 0 }  ][line width=0.08]  [draw opacity=0] (7.2,-1.8) -- (0,0) -- (7.2,1.8) -- cycle    ;
\draw    (248.02,177.46) -- (221.65,195.54) ;
\draw [shift={(220,196.67)}, rotate = 325.57] [fill={rgb, 255:red, 0; green, 0; blue, 0 }  ][line width=0.08]  [draw opacity=0] (7.2,-1.8) -- (0,0) -- (7.2,1.8) -- cycle    ;
\draw [shift={(249.67,176.33)}, rotate = 145.57] [fill={rgb, 255:red, 0; green, 0; blue, 0 }  ][line width=0.08]  [draw opacity=0] (7.2,-1.8) -- (0,0) -- (7.2,1.8) -- cycle    ;
\draw    (285.7,170.78) -- (260.63,184.56) ;
\draw [shift={(258,186)}, rotate = 331.21] [fill={rgb, 255:red, 0; green, 0; blue, 0 }  ][line width=0.08]  [draw opacity=0] (3.57,-1.72) -- (0,0) -- (3.57,1.72) -- cycle    ;
\draw [shift={(288.33,169.33)}, rotate = 151.21] [fill={rgb, 255:red, 0; green, 0; blue, 0 }  ][line width=0.08]  [draw opacity=0] (3.57,-1.72) -- (0,0) -- (3.57,1.72) -- cycle    ;
\draw    (271.84,172.26) -- (277.16,181.41) ;
\draw [shift={(278.67,184)}, rotate = 239.83] [fill={rgb, 255:red, 0; green, 0; blue, 0 }  ][line width=0.08]  [draw opacity=0] (3.57,-1.72) -- (0,0) -- (3.57,1.72) -- cycle    ;
\draw [shift={(270.33,169.67)}, rotate = 59.83] [fill={rgb, 255:red, 0; green, 0; blue, 0 }  ][line width=0.08]  [draw opacity=0] (3.57,-1.72) -- (0,0) -- (3.57,1.72) -- cycle    ;
\draw    (237.33,180.5) -- (241.34,186.5) ;
\draw [shift={(243,189)}, rotate = 236.31] [fill={rgb, 255:red, 0; green, 0; blue, 0 }  ][line width=0.08]  [draw opacity=0] (3.57,-1.72) -- (0,0) -- (3.57,1.72) -- cycle    ;
\draw [shift={(235.67,178)}, rotate = 56.31] [fill={rgb, 255:red, 0; green, 0; blue, 0 }  ][line width=0.08]  [draw opacity=0] (3.57,-1.72) -- (0,0) -- (3.57,1.72) -- cycle    ;
\draw    (292.33,49.17) -- (261.64,65.74) ;
\draw [shift={(259,67.17)}, rotate = 331.63] [fill={rgb, 255:red, 0; green, 0; blue, 0 }  ][line width=0.08]  [draw opacity=0] (5.36,-2.57) -- (0,0) -- (5.36,2.57) -- cycle    ;
\draw    (95.67,189.83) -- (109.33,189.83) ;
\draw [shift={(112.33,189.83)}, rotate = 180] [fill={rgb, 255:red, 0; green, 0; blue, 0 }  ][line width=0.08]  [draw opacity=0] (5.36,-2.57) -- (0,0) -- (5.36,2.57) -- cycle    ;
\draw    (282,88.67) -- (239.43,119.41) ;
\draw [shift={(237,121.17)}, rotate = 324.16] [fill={rgb, 255:red, 0; green, 0; blue, 0 }  ][line width=0.08]  [draw opacity=0] (5.36,-2.57) -- (0,0) -- (5.36,2.57) -- cycle    ;
\draw    (167.67,256.5) -- (252.16,228.12) ;
\draw [shift={(255,227.17)}, rotate = 161.43] [fill={rgb, 255:red, 0; green, 0; blue, 0 }  ][line width=0.08]  [draw opacity=0] (5.36,-2.57) -- (0,0) -- (5.36,2.57) -- cycle    ;

\draw (93,144.23) node [anchor=north west][inner sep=0.75pt]  [font=\small]  {$p_{1}$};
\draw (98,90) node [anchor=north west][inner sep=0.75pt]  [font=\small]  {$p_{2}$};
\draw (129,226) node [anchor=north west][inner sep=0.75pt]  [font=\small]  {$p_{3}$};
\draw (214.33,180) node [anchor=north west][inner sep=0.75pt]  [font=\small]  {$p_{4}$};
\draw (240.62,181) node [anchor=north west][inner sep=0.75pt]  [font=\small]  {$p_{5}$};
\draw (287.67,162.23) node [anchor=north west][inner sep=0.75pt]  [font=\small]  {$p_{6}$};
\draw (260.5,160) node [anchor=north west][inner sep=0.75pt]  [font=\small]  {$p_{7}$};
\draw (311.5,43.42) node   [align=left] {\begin{minipage}[lt]{30.15pt}\setlength\topsep{0pt}
stator
\end{minipage}};
\draw (85,188) node   [align=left] {\begin{minipage}[lt]{25.61pt}\setlength\topsep{0pt}
rotor
\end{minipage}};
\draw (320,86) node   [align=left] {\begin{minipage}[lt]{53.27pt}\setlength\topsep{0pt}
stator slot
\end{minipage}};
\draw (149,255) node  [font=\small] [align=left] {\begin{minipage}[lt]{40.12pt}\setlength\topsep{0pt}
rotor slot
\end{minipage}};

\end{tikzpicture}}

%% file: fig/psm_technology_sample1.tex
\tikzset{every picture/.style={line width=0.75pt}} 
\resizebox{\linewidth}{!}{
\begin{tikzpicture}[x=0.75pt,y=0.75pt,yscale=-1,xscale=1]

\draw (246.53,175.76) node  {\includegraphics[width=195.86pt,height=189.64pt]{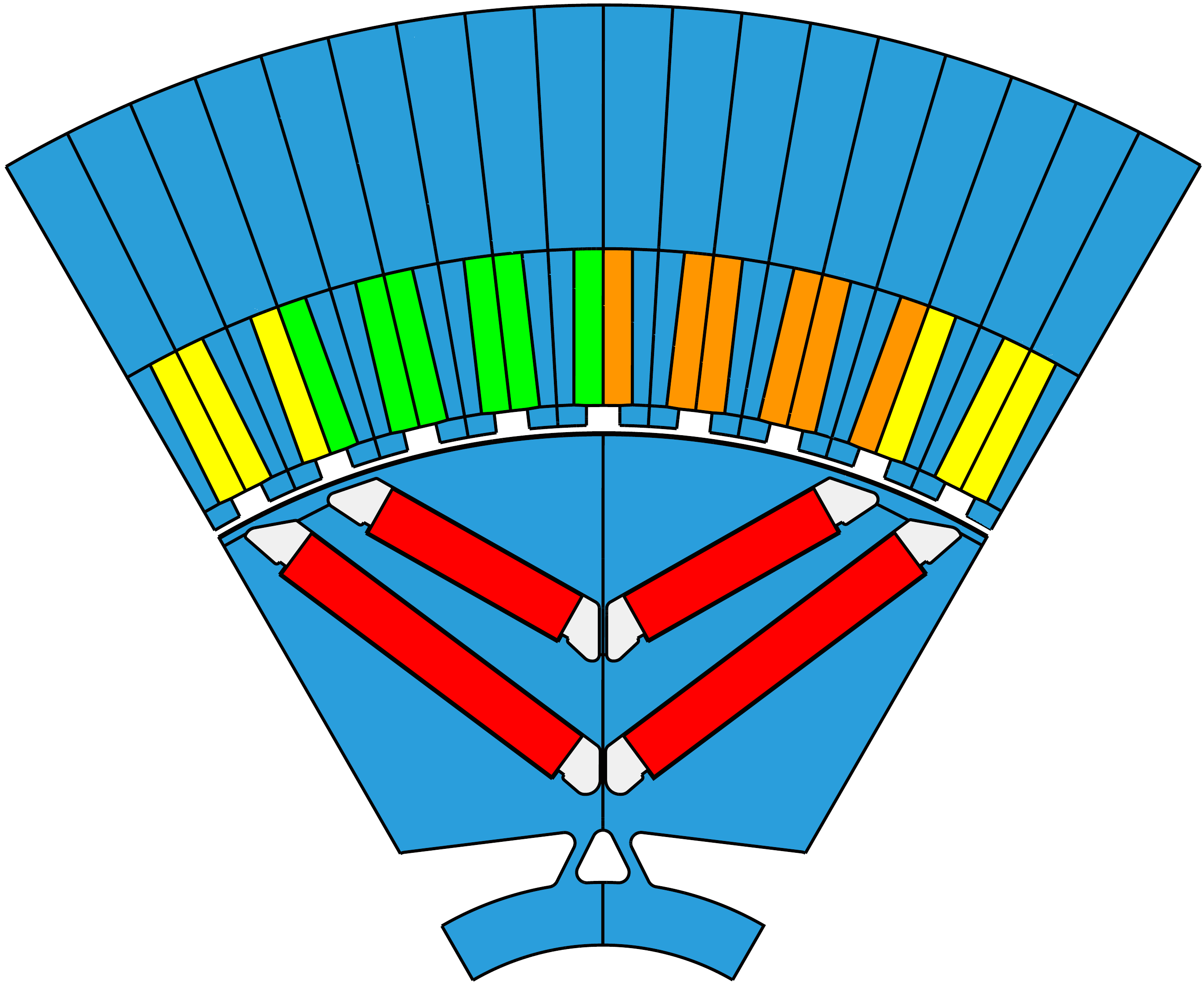}};
\draw    (142.97,148.94) -- (159.09,182.11) ;
\draw [shift={(159.96,183.91)}, rotate = 244.08] [fill={rgb, 255:red, 0; green, 0; blue, 0 }  ][line width=0.08]  [draw opacity=0] (4.8,-1.2) -- (0,0) -- (4.8,1.2) -- cycle    ;
\draw [shift={(142.1,147.14)}, rotate = 64.08] [fill={rgb, 255:red, 0; green, 0; blue, 0 }  ][line width=0.08]  [draw opacity=0] (4.8,-1.2) -- (0,0) -- (4.8,1.2) -- cycle    ;
\draw    (161.16,186.66) -- (162.1,188.82) ;
\draw [shift={(163.29,191.57)}, rotate = 246.5] [fill={rgb, 255:red, 0; green, 0; blue, 0 }  ][line width=0.08]  [draw opacity=0] (3.57,-1.72) -- (0,0) -- (3.57,1.72) -- cycle    ;
\draw [shift={(159.96,183.91)}, rotate = 66.5] [fill={rgb, 255:red, 0; green, 0; blue, 0 }  ][line width=0.08]  [draw opacity=0] (3.57,-1.72) -- (0,0) -- (3.57,1.72) -- cycle    ;
\draw    (220.02,190.12) -- (215.8,196.47) ;
\draw [shift={(214.14,198.96)}, rotate = 303.59] [fill={rgb, 255:red, 0; green, 0; blue, 0 }  ][line width=0.08]  [draw opacity=0] (3.57,-1.72) -- (0,0) -- (3.57,1.72) -- cycle    ;
\draw [shift={(221.67,187.62)}, rotate = 123.59] [fill={rgb, 255:red, 0; green, 0; blue, 0 }  ][line width=0.08]  [draw opacity=0] (3.57,-1.72) -- (0,0) -- (3.57,1.72) -- cycle    ;
\draw  [dash pattern={on 0.84pt off 2.51pt}]  (260.99,170.85) -- (291.33,173.67) ;
\draw [shift={(259,170.67)}, rotate = 5.3] [fill={rgb, 255:red, 0; green, 0; blue, 0 }  ][line width=0.08]  [draw opacity=0] (4.8,-1.2) -- (0,0) -- (4.8,1.2) -- cycle    ;
\draw  [draw opacity=0] (281.31,179.23) .. controls (280.95,179.3) and (280.56,179.33) .. (280.17,179.33) .. controls (277.68,179.33) and (275.67,177.88) .. (275.67,176.08) .. controls (275.67,174.29) and (277.68,172.83) .. (280.17,172.83) .. controls (280.56,172.83) and (280.95,172.87) .. (281.31,172.94) -- (280.17,176.08) -- cycle ; \draw   (281.31,179.23) .. controls (280.95,179.3) and (280.56,179.33) .. (280.17,179.33) .. controls (277.68,179.33) and (275.67,177.88) .. (275.67,176.08) .. controls (275.67,174.29) and (277.68,172.83) .. (280.17,172.83) .. controls (280.56,172.83) and (280.95,172.87) .. (281.31,172.94) ;  
\draw    (184.02,170.37) -- (174.93,174.92) ;
\draw [shift={(173.14,175.81)}, rotate = 333.43] [fill={rgb, 255:red, 0; green, 0; blue, 0 }  ][line width=0.08]  [draw opacity=0] (4.8,-1.2) -- (0,0) -- (4.8,1.2) -- cycle    ;
\draw [shift={(185.81,169.48)}, rotate = 153.43] [fill={rgb, 255:red, 0; green, 0; blue, 0 }  ][line width=0.08]  [draw opacity=0] (4.8,-1.2) -- (0,0) -- (4.8,1.2) -- cycle    ;
\draw    (111.86,77.11) -- (135.01,106.74) ;
\draw [shift={(136.86,109.11)}, rotate = 232] [fill={rgb, 255:red, 0; green, 0; blue, 0 }  ][line width=0.08]  [draw opacity=0] (5.36,-2.57) -- (0,0) -- (5.36,2.57) -- cycle    ;
\draw    (327.21,54.04) -- (323.61,60.88) ;
\draw [shift={(322.21,63.54)}, rotate = 297.76] [fill={rgb, 255:red, 0; green, 0; blue, 0 }  ][line width=0.08]  [draw opacity=0] (5.36,-2.57) -- (0,0) -- (5.36,2.57) -- cycle    ;
\draw    (319.5,233.75) -- (303.76,230.75) ;
\draw [shift={(300.82,230.19)}, rotate = 10.78] [fill={rgb, 255:red, 0; green, 0; blue, 0 }  ][line width=0.08]  [draw opacity=0] (5.36,-2.57) -- (0,0) -- (5.36,2.57) -- cycle    ;
\draw    (345.96,155.34) -- (313.74,140.97) ;
\draw [shift={(311,139.75)}, rotate = 24.03] [fill={rgb, 255:red, 0; green, 0; blue, 0 }  ][line width=0.08]  [draw opacity=0] (5.36,-2.57) -- (0,0) -- (5.36,2.57) -- cycle    ;
\draw    (292.6,241.7) -- (274.62,227.75) ;
\draw [shift={(272.25,225.91)}, rotate = 37.81] [fill={rgb, 255:red, 0; green, 0; blue, 0 }  ][line width=0.08]  [draw opacity=0] (5.36,-2.57) -- (0,0) -- (5.36,2.57) -- cycle    ;
\draw    (323.8,213.3) -- (287.47,188.35) ;
\draw [shift={(285,186.65)}, rotate = 34.48] [fill={rgb, 255:red, 0; green, 0; blue, 0 }  ][line width=0.08]  [draw opacity=0] (5.36,-2.57) -- (0,0) -- (5.36,2.57) -- cycle    ;

\draw (144,184.35) node [anchor=north west][inner sep=0.75pt]  [font=\normalsize]  {$p_{3}$};
\draw (132,158.38) node [anchor=north west][inner sep=0.75pt]  [font=\normalsize]  {$p_{4}$};
\draw (259.17,172) node [anchor=north west][inner sep=0.75pt]  [font=\normalsize]  {$p_{5}$};
\draw (168.29,160) node [anchor=north west][inner sep=0.75pt]  [font=\normalsize]  {$p_{7}$};
\draw (214.5,175.07) node [anchor=north west][inner sep=0.75pt]  [font=\normalsize]  {$p_{6}$};
\draw (231.36,161.5) node [anchor=north west][inner sep=0.75pt]  [font=\normalsize]  {$p_{2}$};
\draw (106.46,71.02) node  [font=\normalsize,rotate=-308.84] [align=left] {\begin{minipage}[lt]{52.7pt}\setlength\topsep{0pt}
stator yoke
\end{minipage}};
\draw (334.85,46.25) node  [font=\normalsize,rotate=-0.6] [align=left] {\begin{minipage}[lt]{27.28pt}\setlength\topsep{0pt}
stator 
\end{minipage}};
\draw (337.06,235) node  [font=\normalsize,rotate=-0.6] [align=left] {\begin{minipage}[lt]{27.28pt}\setlength\topsep{0pt}
rotor 
\end{minipage}};
\draw (360.26,142.65) node  [font=\small,rotate=-295.48] [align=left] {\begin{minipage}[lt]{87.49pt}\setlength\topsep{0pt}
stator tooth winding 
\end{minipage}};
\draw (221.47,51.55) node [anchor=north west][inner sep=0.75pt]  [font=\normalsize]  {$p_{1}$};
\draw (352.2,215) node  [font=\normalsize,rotate=-0.6] [align=left] {\begin{minipage}[lt]{45.76pt}\setlength\topsep{0pt}
magnet\\layer1 
\end{minipage}};
\draw (309,254) node  [font=\normalsize,rotate=-0.6] [align=left] {\begin{minipage}[lt]{45.76pt}\setlength\topsep{0pt}
magnet\\layer2 
\end{minipage}};

\end{tikzpicture}}

%% file: fig/flow_chart_dataset_generation.tex
        \tikzstyle{block} = [rectangle, draw, fill=black!10, 
            text width=12em, text centered, rounded corners,
            minimum height=3em]
        \tikzstyle{block1} = [rectangle, draw, fill=black!10,
            text width=6em, text centered, rounded corners,
            minimum height=3em]
        \tikzstyle{line} = [draw, -latex']
        \begin{tikzpicture}[node distance=4.5em]
            \tikzset{font=\footnotesize}
            \node [block] (init) {Define input parameters $\mathbf{p}$ with limits, e.g., $p_{i}$ from \autoref{tab:input_asm} or \autoref{tab:input_psm}};
            \node [block, below of=init] (lhs) {Build initial population $\mathcal{D}_{t}$ using LHS};
            \node [block, below of=lhs] (cad) {Perform geometry check and filter erroneous designs};
            \node [block, below of=cad] (pareto) {Compute KPIs $\mathbf{k}$ via magneto-static FE simulation or analytical calculation };
            \node [block, below of=pareto] (data) {Collect the data and store it in the database };
         
            \path [line] (init) -- (lhs);
            \path [line] (lhs) -- (cad);
            \path [line] (cad) -- (pareto);
            \path [line] (pareto) -- (data);

        \end{tikzpicture}

%% file: fig/asm_param_distr.tex
\tikzset{every picture/.style={line width=0.75pt}} 
\resizebox{0.8\linewidth}{!}{
\begin{tikzpicture}[x=0.75pt,y=0.75pt,yscale=-1,xscale=1]
 \tikzstyle{every node}=[font=\normalsize]

\draw (256.63,153.2) node  {\includegraphics[width=221.48pt,height=211.2pt]{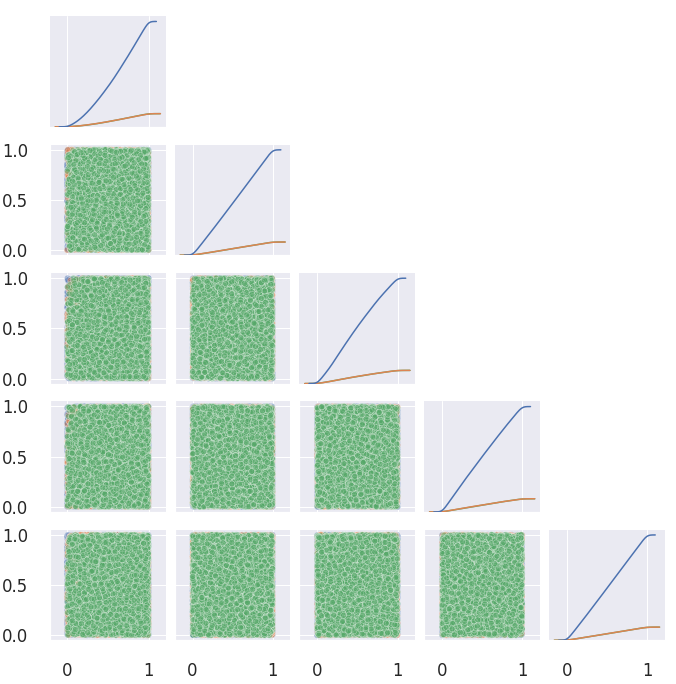}};
\draw  [fill={rgb, 255:red, 74; green, 144; blue, 226 }  ,fill opacity=1 ] (337.43,17.18) .. controls (337.43,15.34) and (338.85,13.85) .. (340.59,13.85) .. controls (342.34,13.85) and (343.75,15.34) .. (343.75,17.18) .. controls (343.75,19.01) and (342.34,20.5) .. (340.59,20.5) .. controls (338.85,20.5) and (337.43,19.01) .. (337.43,17.18) -- cycle ;
\draw  [fill={rgb, 255:red, 239; green, 181; blue, 84 }  ,fill opacity=1 ] (337.36,32.95) .. controls (337.36,31.12) and (338.77,29.63) .. (340.52,29.63) .. controls (342.26,29.63) and (343.68,31.12) .. (343.68,32.95) .. controls (343.68,34.79) and (342.26,36.28) .. (340.52,36.28) .. controls (338.77,36.28) and (337.36,34.79) .. (337.36,32.95) -- cycle ;
\draw  [fill={rgb, 255:red, 126; green, 211; blue, 33 }  ,fill opacity=1 ] (337.24,48.4) .. controls (337.24,46.56) and (338.66,45.07) .. (340.4,45.07) .. controls (342.15,45.07) and (343.56,46.56) .. (343.56,48.4) .. controls (343.56,50.23) and (342.15,51.72) .. (340.4,51.72) .. controls (338.66,51.72) and (337.24,50.23) .. (337.24,48.4) -- cycle ;

\draw (348.64,11) node [anchor=north west][inner sep=0.75pt]  [font=\normalsize] [align=left] {Train};
\draw (348.92,29.14) node [anchor=north west][inner sep=0.75pt]  [font=\normalsize] [align=left] {Validation};
\draw (348.92,44.57) node [anchor=north west][inner sep=0.75pt]  [font=\normalsize] [align=left] {Test};
\draw (151.53,290.52) node [anchor=north west][inner sep=0.75pt]  [font=\normalsize]  {$p_{1}$};
\draw (203.65,290.38) node [anchor=north west][inner sep=0.75pt]  [font=\normalsize]  {$p_{2}$};
\draw (258.22,290.16) node [anchor=north west][inner sep=0.75pt]  [font=\normalsize]  {$p_{3}$};
\draw (311.76,290.64) node [anchor=north west][inner sep=0.75pt]  [font=\normalsize]  {$p_{4}$};
\draw (367.37,291.21) node [anchor=north west][inner sep=0.75pt]  [font=\normalsize]  {$p_{5}$};
\draw (90.76,87.19) node [anchor=north west][inner sep=0.75pt]  [font=\normalsize]  {$p_{2}$};
\draw (90.94,140.35) node [anchor=north west][inner sep=0.75pt]  [font=\normalsize]  {$p_{3}$};
\draw (90.06,194.28) node [anchor=north west][inner sep=0.75pt]  [font=\normalsize]  {$p_{4}$};
\draw (90.27,246.59) node [anchor=north west][inner sep=0.75pt]  [font=\normalsize]  {$p_{5}$};

\end{tikzpicture}}

%% file: fig/psm_param_distr.tex
\tikzset{every picture/.style={line width=0.75pt}} 
\resizebox{0.8\linewidth}{!}{
\begin{tikzpicture}[x=0.75pt,y=0.75pt,yscale=-1,xscale=1]
 \tikzstyle{every node}=[font=\normalsize]

\draw (237.15,149.39) node  {\includegraphics[width=222.25pt,height=210.92pt]{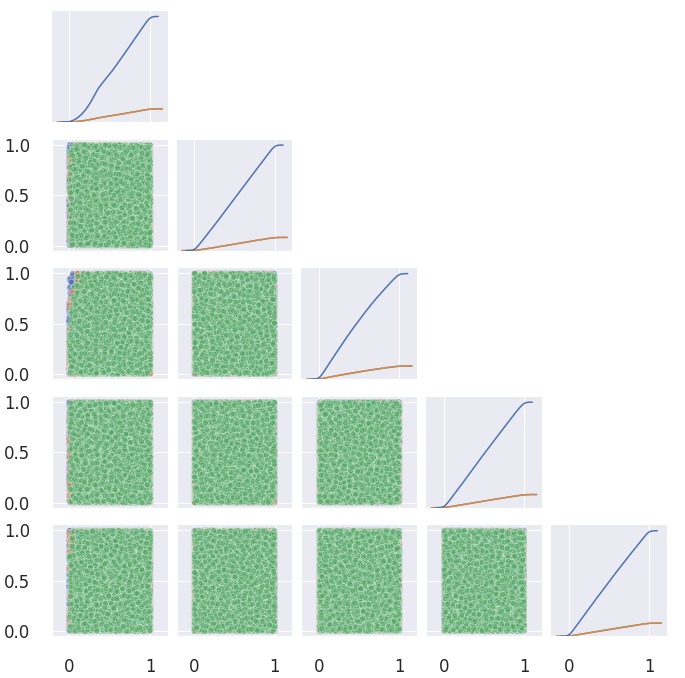}};
\draw  [fill={rgb, 255:red, 74; green, 144; blue, 226 }  ,fill opacity=1 ] (319.77,19.18) .. controls (319.77,17.34) and (321.18,15.85) .. (322.93,15.85) .. controls (324.67,15.85) and (326.09,17.34) .. (326.09,19.18) .. controls (326.09,21.01) and (324.67,22.5) .. (322.93,22.5) .. controls (321.18,22.5) and (319.77,21.01) .. (319.77,19.18) -- cycle ;
\draw  [fill={rgb, 255:red, 239; green, 181; blue, 84 }  ,fill opacity=1 ] (319.69,34.95) .. controls (319.69,33.12) and (321.11,31.63) .. (322.85,31.63) .. controls (324.6,31.63) and (326.01,33.12) .. (326.01,34.95) .. controls (326.01,36.79) and (324.6,38.28) .. (322.85,38.28) .. controls (321.11,38.28) and (319.69,36.79) .. (319.69,34.95) -- cycle ;
\draw  [fill={rgb, 255:red, 126; green, 211; blue, 33 }  ,fill opacity=1 ] (319.58,50.4) .. controls (319.58,48.56) and (320.99,47.07) .. (322.74,47.07) .. controls (324.48,47.07) and (325.9,48.56) .. (325.9,50.4) .. controls (325.9,52.23) and (324.48,53.72) .. (322.74,53.72) .. controls (320.99,53.72) and (319.58,52.23) .. (319.58,50.4) -- cycle ;

\draw (330.97,13) node [anchor=north west][inner sep=0.75pt]  [font=\normalsize] [align=left] {Train};
\draw (331.25,31.14) node [anchor=north west][inner sep=0.75pt]  [font=\normalsize] [align=left] {Validation};
\draw (331.25,46.57) node [anchor=north west][inner sep=0.75pt]  [font=\normalsize] [align=left] {Test};
\draw (131.72,286.52) node [anchor=north west][inner sep=0.75pt]  [font=\normalsize]  {$p_{1}$};
\draw (183.84,286.38) node [anchor=north west][inner sep=0.75pt]  [font=\normalsize]  {$p_{2}$};
\draw (238.41,286.16) node [anchor=north west][inner sep=0.75pt]  [font=\normalsize]  {$p_{3}$};
\draw (291.95,286.64) node [anchor=north west][inner sep=0.75pt]  [font=\normalsize]  {$p_{4}$};
\draw (347.56,287.21) node [anchor=north west][inner sep=0.75pt]  [font=\normalsize]  {$p_{5}$};
\draw (70.95,83.19) node [anchor=north west][inner sep=0.75pt]  [font=\normalsize]  {$p_{2}$};
\draw (71.13,136.35) node [anchor=north west][inner sep=0.75pt]  [font=\normalsize]  {$p_{3}$};
\draw (70.25,190.28) node [anchor=north west][inner sep=0.75pt]  [font=\normalsize]  {$p_{4}$};
\draw (72.46,242.59) node [anchor=north west][inner sep=0.75pt]  [font=\normalsize]  {$p_{5}$};

\end{tikzpicture}}

%% file: fig/asm_kpi_distr.tex
\tikzset{every picture/.style={line width=0.75pt}} 
\resizebox{0.8\linewidth}{!}{
\begin{tikzpicture}[x=0.75pt,y=0.75pt,yscale=-1,xscale=1]
 \tikzstyle{every node}=[font=\normalsize]


\draw (276.03,148.71) node  {\includegraphics[width=220.58pt,height=211.93pt]{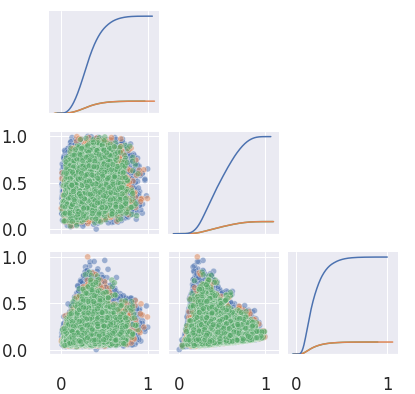}};
\draw  [fill={rgb, 255:red, 74; green, 144; blue, 226 }  ,fill opacity=1 ] (356.43,18.58) .. controls (356.43,16.74) and (357.85,15.25) .. (359.59,15.25) .. controls (361.34,15.25) and (362.75,16.74) .. (362.75,18.58) .. controls (362.75,20.41) and (361.34,21.9) .. (359.59,21.9) .. controls (357.85,21.9) and (356.43,20.41) .. (356.43,18.58) -- cycle ;
\draw  [fill={rgb, 255:red, 239; green, 181; blue, 84 }  ,fill opacity=1 ] (356.36,34.35) .. controls (356.36,32.52) and (357.77,31.03) .. (359.52,31.03) .. controls (361.26,31.03) and (362.68,32.52) .. (362.68,34.35) .. controls (362.68,36.19) and (361.26,37.68) .. (359.52,37.68) .. controls (357.77,37.68) and (356.36,36.19) .. (356.36,34.35) -- cycle ;
\draw  [fill={rgb, 255:red, 126; green, 211; blue, 33 }  ,fill opacity=1 ] (356.24,49.8) .. controls (356.24,47.96) and (357.66,46.47) .. (359.4,46.47) .. controls (361.15,46.47) and (362.56,47.96) .. (362.56,49.8) .. controls (362.56,51.63) and (361.15,53.12) .. (359.4,53.12) .. controls (357.66,53.12) and (356.24,51.63) .. (356.24,49.8) -- cycle ;

\draw (367.64,12.4) node [anchor=north west][inner sep=0.75pt]  [font=\normalsize] [align=left] {Train};
\draw (367.92,30.54) node [anchor=north west][inner sep=0.75pt]  [font=\normalsize] [align=left] {Validation};
\draw (367.92,45.97) node [anchor=north west][inner sep=0.75pt]  [font=\normalsize] [align=left] {Test};
\draw (103.68,131.4) node [anchor=north west][inner sep=0.75pt]    {$k_{2}$};
\draw (104.35,217.07) node [anchor=north west][inner sep=0.75pt]    {$k_{3}$};
\draw (198.35,288.4) node [anchor=north west][inner sep=0.75pt]    {$k_{1}$};
\draw (282.35,287.33) node [anchor=north west][inner sep=0.75pt]    {$k_{2}$};
\draw (367.01,287) node [anchor=north west][inner sep=0.75pt]    {$k_{3}$};

\end{tikzpicture}}

%% file: fig/psm_kpi_distr.tex
\tikzset{every picture/.style={line width=0.75pt}} 
\resizebox{0.8\linewidth}{!}{
\begin{tikzpicture}[x=0.75pt,y=0.75pt,yscale=-1,xscale=1]
 \tikzstyle{every node}=[font=\normalsize]

\draw (216.36,151.14) node  {\includegraphics[width=221.08pt,height=209.89pt]{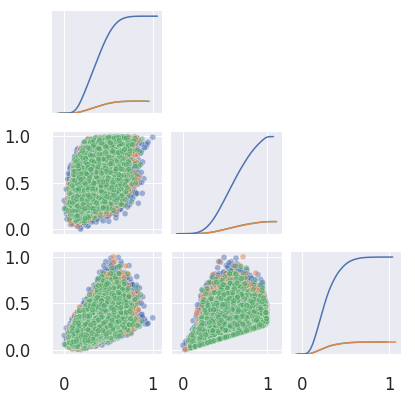}};
\draw  [fill={rgb, 255:red, 74; green, 144; blue, 226 }  ,fill opacity=1 ] (297.23,22.83) .. controls (297.23,20.99) and (298.65,19.5) .. (300.39,19.5) .. controls (302.14,19.5) and (303.55,20.99) .. (303.55,22.83) .. controls (303.55,24.66) and (302.14,26.15) .. (300.39,26.15) .. controls (298.65,26.15) and (297.23,24.66) .. (297.23,22.83) -- cycle ;
\draw  [fill={rgb, 255:red, 239; green, 181; blue, 84 }  ,fill opacity=1 ] (297.16,38.6) .. controls (297.16,36.77) and (298.57,35.28) .. (300.32,35.28) .. controls (302.06,35.28) and (303.48,36.77) .. (303.48,38.6) .. controls (303.48,40.44) and (302.06,41.93) .. (300.32,41.93) .. controls (298.57,41.93) and (297.16,40.44) .. (297.16,38.6) -- cycle ;
\draw  [fill={rgb, 255:red, 126; green, 211; blue, 33 }  ,fill opacity=1 ] (297.04,54.05) .. controls (297.04,52.21) and (298.46,50.72) .. (300.2,50.72) .. controls (301.95,50.72) and (303.36,52.21) .. (303.36,54.05) .. controls (303.36,55.88) and (301.95,57.37) .. (300.2,57.37) .. controls (298.46,57.37) and (297.04,55.88) .. (297.04,54.05) -- cycle ;

\draw (308.44,16.65) node [anchor=north west][inner sep=0.75pt]  [font=\normalsize] [align=left] {Train};
\draw (308.72,34.79) node [anchor=north west][inner sep=0.75pt]  [font=\normalsize] [align=left] {Validation};
\draw (308.72,50.22) node [anchor=north west][inner sep=0.75pt]  [font=\normalsize] [align=left] {Test};
\draw (47.68,131.4) node [anchor=north west][inner sep=0.75pt]    {$k_{2}$};
\draw (48.35,217.07) node [anchor=north west][inner sep=0.75pt]    {$k_{3}$};
\draw (142.35,288.4) node [anchor=north west][inner sep=0.75pt]    {$k_{1}$};
\draw (226.35,287.33) node [anchor=north west][inner sep=0.75pt]    {$k_{2}$};
\draw (311.01,287) node [anchor=north west][inner sep=0.75pt]    {$k_{3}$};

\end{tikzpicture}}

%% file: fig/workflow_training_vae.tex
\tikzset{every picture/.style={line width=0.75pt}} 
\resizebox{\linewidth}{!}{
\begin{tikzpicture}[x=0.75pt,y=0.75pt,yscale=-1,xscale=1]

\draw (128,188) node  {\includegraphics[width=30pt,height=30pt]{fig/Pareto_VAE_PSM1.pdf}};
\draw (129,126) node  {\includegraphics[width=30pt,height=30pt]{fig/Pareto_VAE_ASM1.pdf}};
\draw   (96.4,88.1) -- (157,88.1) -- (157,212.05) -- (96.4,212.05) -- cycle ;
\draw    (95.8,150.9) -- (157.25,150.67) ;
\draw    (156.58,150.77) -- (166.84,150.61) ;
\draw [shift={(169.84,150.56)}, rotate = 179.09] [fill={rgb, 255:red, 0; green, 0; blue, 0 }  ][line width=0.08]  [draw opacity=0] (3.57,-1.72) -- (0,0) -- (3.57,1.72) -- cycle    ;
\draw  [fill={rgb, 255:red, 155; green, 155; blue, 155 }  ,fill opacity=1 ] (170.04,88.81) -- (177,88.81) -- (177,212.76) -- (170.04,212.76) -- cycle ;
\draw   (179.54,89.66) -- (237.02,106.91) -- (237.02,195) -- (179.54,212.24) -- cycle ;
\draw  [fill={rgb, 255:red, 155; green, 155; blue, 155 }  ,fill opacity=1 ] (240.34,107.49) -- (254.2,107.49) -- (254.2,194.91) -- (240.34,194.91) -- cycle ;
\draw    (239.63,150.74) -- (254.2,150.74) ;
\draw    (254.53,130.1) -- (282.1,129.97) -- (282.82,129.98) ;
\draw [shift={(285.82,130.04)}, rotate = 181.04] [fill={rgb, 255:red, 0; green, 0; blue, 0 }  ][line width=0.08]  [draw opacity=0] (3.57,-1.72) -- (0,0) -- (3.57,1.72) -- cycle    ;
\draw    (254.53,169.77) -- (282.1,169.64) -- (282.82,169.65) ;
\draw [shift={(285.82,169.7)}, rotate = 181.04] [fill={rgb, 255:red, 0; green, 0; blue, 0 }  ][line width=0.08]  [draw opacity=0] (3.57,-1.72) -- (0,0) -- (3.57,1.72) -- cycle    ;
\draw   (285.82,130.04) .. controls (285.89,125.87) and (289.12,122.55) .. (293.04,122.62) .. controls (296.95,122.69) and (300.07,126.12) .. (300,130.29) .. controls (299.92,134.45) and (296.69,137.77) .. (292.78,137.71) .. controls (288.86,137.64) and (285.75,134.2) .. (285.82,130.04) -- cycle ; \draw   (285.82,130.04) -- (300,130.29) ; \draw   (293.04,122.62) -- (292.78,137.71) ;
\draw   (285.82,169.7) .. controls (285.82,165.96) and (288.85,162.93) .. (292.6,162.93) .. controls (296.34,162.93) and (299.37,165.96) .. (299.37,169.7) .. controls (299.37,173.45) and (296.34,176.48) .. (292.6,176.48) .. controls (288.85,176.48) and (285.82,173.45) .. (285.82,169.7) -- cycle ;
\draw  [fill={rgb, 255:red, 0; green, 0; blue, 0 }  ,fill opacity=1 ] (291.18,169.7) .. controls (291.18,168.92) and (291.81,168.29) .. (292.6,168.29) .. controls (293.38,168.29) and (294.01,168.92) .. (294.01,169.7) .. controls (294.01,170.49) and (293.38,171.12) .. (292.6,171.12) .. controls (291.81,171.12) and (291.18,170.49) .. (291.18,169.7) -- cycle ;
\draw    (292.6,162.93) -- (292.75,140.71) ;
\draw [shift={(292.78,137.71)}, rotate = 90.41] [fill={rgb, 255:red, 0; green, 0; blue, 0 }  ][line width=0.08]  [draw opacity=0] (3.57,-1.72) -- (0,0) -- (3.57,1.72) -- cycle    ;
\draw    (300,130.29) -- (318.9,130.16) -- (319.61,130.17) ;
\draw [shift={(322.61,130.22)}, rotate = 181.04] [fill={rgb, 255:red, 0; green, 0; blue, 0 }  ][line width=0.08]  [draw opacity=0] (3.57,-1.72) -- (0,0) -- (3.57,1.72) -- cycle    ;
\draw  [fill={rgb, 255:red, 155; green, 155; blue, 155 }  ,fill opacity=1 ] (322.01,107.43) -- (335.87,107.43) -- (335.87,193.1) -- (322.01,193.1) -- cycle ;
\draw    (330.2,89.1) -- (330.2,104.1) ;
\draw [shift={(330.2,107.1)}, rotate = 270] [fill={rgb, 255:red, 0; green, 0; blue, 0 }  ][line width=0.08]  [draw opacity=0] (3.57,-1.72) -- (0,0) -- (3.57,1.72) -- cycle    ;
\draw   (397.02,213.71) -- (339.54,196.46) -- (339.54,108.37) -- (397.02,91.13) -- cycle ;
\draw   (340.87,215.33) -- (398.36,232.57) -- (398.36,320.67) -- (340.87,337.91) -- cycle ;
\draw    (329.87,193.77) -- (329.67,259.57) -- (338.49,259.92) ;
\draw [shift={(341.49,260.04)}, rotate = 182.28] [fill={rgb, 255:red, 0; green, 0; blue, 0 }  ][line width=0.08]  [draw opacity=0] (3.57,-1.72) -- (0,0) -- (3.57,1.72) -- cycle    ;
\draw  [fill={rgb, 255:red, 155; green, 155; blue, 155 }  ,fill opacity=1 ] (400.68,91.43) -- (406.37,91.43) -- (406.37,213.7) -- (400.68,213.7) -- cycle ;
\draw  [fill={rgb, 255:red, 155; green, 155; blue, 155 }  ,fill opacity=1 ] (400.68,233.77) -- (405.8,233.77) -- (405.8,320.63) -- (400.68,320.63) -- cycle ;
\draw    (292.67,192.7) -- (292.61,179.48) ;
\draw [shift={(292.6,176.48)}, rotate = 89.75] [fill={rgb, 255:red, 0; green, 0; blue, 0 }  ][line width=0.08]  [draw opacity=0] (3.57,-1.72) -- (0,0) -- (3.57,1.72) -- cycle    ;

\draw (261.67,107.2) node [anchor=north west][inner sep=0.75pt]  [font=\normalsize] [align=left] {sampling};
\draw (301,75.87) node [anchor=north west][inner sep=0.75pt]  [font=\normalsize] [align=left] {latent vector};
\draw (181.33,122.53) node [anchor=north west][inner sep=0.75pt]  [font=\normalsize] [align=left] {encoder \\network};
\draw (342.67,123.87) node [anchor=north west][inner sep=0.75pt]  [font=\normalsize] [align=left] {decoder \\network};
\draw (342,249.53) node [anchor=north west][inner sep=0.75pt]  [font=\normalsize] [align=left] {KPIs \\predictior};
\draw (168.44,75) node [anchor=north west][inner sep=0.75pt]    {$\mathbf{p}$};
\draw (241.2,159.87) node [anchor=north west][inner sep=0.75pt]    {$\bm{\sigma }$};
\draw (241.2,121.53) node [anchor=north west][inner sep=0.75pt]    {$\bm{\upsilon }$};
\draw (248.68,195.98) node [anchor=north west][inner sep=0.75pt]  [font=\normalsize]  {$\varepsilon \sim \mathcal{N}( 0,I)$};
\draw (399.3,75) node [anchor=north west][inner sep=0.75pt]    {$\hat{\mathbf{p}}$};
\draw (397.37,321.17) node [anchor=north west][inner sep=0.75pt]    {$\hat{\mathbf{k}}$};
\draw (108,155.2) node [anchor=north west][inner sep=0.75pt]  [font=\small] [align=left] {PMSM};
\draw (109,91.2) node [anchor=north west][inner sep=0.75pt]  [font=\small] [align=left] {ASM};
\draw (323.2,137.53) node [anchor=north west][inner sep=0.75pt]    {$\mathbf{z}$};

\end{tikzpicture}}

%% file: fig/vae_optimization_workflow.tex
\tikzset{every picture/.style={line width=0.75pt}} 
\resizebox{\linewidth}{!}{

\begin{tikzpicture}[x=0.75pt,y=0.75pt,yscale=-1,xscale=1]

\draw   (304.67,95.5) -- (365,113.6) -- (365,187.4) -- (304.67,205.5) -- cycle ;
\draw   (146.33,206.5) -- (86,188.4) -- (86,114.6) -- (146.33,96.5) -- cycle ;
\draw  [fill={rgb, 255:red, 155; green, 155; blue, 155 }  ,fill opacity=1 ] (73.67,114.5) -- (83.33,114.5) -- (83.33,188.17) -- (73.67,188.17) -- cycle ;
\draw  [color={rgb, 255:red, 0; green, 0; blue, 0 }  ,draw opacity=1 ][fill={rgb, 255:red, 248; green, 231; blue, 28 }  ,fill opacity=1 ] (149,96.17) -- (159.67,96.17) -- (159.67,206.83) -- (149,206.83) -- cycle ;
\draw  [fill={rgb, 255:red, 155; green, 155; blue, 155 }  ,fill opacity=1 ] (295,263.5) -- (207.4,263.5) -- (149,250.33) -- (207.4,237.17) -- (295,237.17) -- cycle ;
\draw  [fill={rgb, 255:red, 126; green, 211; blue, 33 }  ,fill opacity=1 ] (367.67,114.5) -- (376.67,114.5) -- (376.67,187.83) -- (367.67,187.83) -- cycle ;
\draw   (229,96.5) -- (289.33,114.6) -- (289.33,188.4) -- (229,206.5) -- cycle ;
\draw  [fill={rgb, 255:red, 74; green, 144; blue, 226 }  ,fill opacity=1 ] (215.67,96.5) -- (225.33,96.5) -- (225.33,207.17) -- (215.67,207.17) -- cycle ;
\draw    (372.33,187.83) -- (371.67,241.5) -- (371.67,250.17) -- (298.67,250.49) ;
\draw [shift={(295.67,250.5)}, rotate = 359.75] [fill={rgb, 255:red, 0; green, 0; blue, 0 }  ][line width=0.08]  [draw opacity=0] (5.36,-2.57) -- (0,0) -- (5.36,2.57) -- cycle    ;
\draw    (149,250.33) -- (78,251.17) -- (78.63,191.5) ;
\draw [shift={(78.67,188.5)}, rotate = 90.61] [fill={rgb, 255:red, 0; green, 0; blue, 0 }  ][line width=0.08]  [draw opacity=0] (5.36,-2.57) -- (0,0) -- (5.36,2.57) -- cycle    ;
\draw [line width=0.75]    (159.67,151.83) -- (213,151.83) ;
\draw [shift={(216,151.83)}, rotate = 180] [fill={rgb, 255:red, 0; green, 0; blue, 0 }  ][line width=0.08]  [draw opacity=0] (5.36,-2.57) -- (0,0) -- (5.36,2.57) -- cycle    ;
\draw  [fill={rgb, 255:red, 155; green, 155; blue, 155 }  ,fill opacity=1 ] (291.67,114.5) -- (301.33,114.5) -- (301.33,188.17) -- (291.67,188.17) -- cycle ;
\draw    (177,128) -- (192.67,128) ;
\draw [shift={(195.67,128.08)}, rotate = 180] [fill={rgb, 255:red, 0; green, 0; blue, 0 }  ][line width=0.08]  [draw opacity=0] (5.36,-2.57) -- (0,0) -- (5.36,2.57) -- cycle    ;

\draw (150,80) node [anchor=north west][inner sep=0.75pt]  [font=\normalsize]  {$\hat{\mathbf{p}}$};
\draw (368,98) node [anchor=north west][inner sep=0.75pt]  [font=\normalsize]  {$\hat{\mathbf{k}}$};
\draw (74,105.4) node [anchor=north west][inner sep=0.75pt]  [font=\normalsize]  {$\mathbf{z}$};
\draw (287,103) node [anchor=north west][inner sep=0.75pt]  [font=\normalsize]  {$\mathbf{z}_{\textrm o}$};
\draw (196,119) node [anchor=north west][inner sep=0.75pt]  [font=\normalsize]  {$\hat{\mathbf{p}}_{\textrm o}$};
\draw (163,119) node [anchor=north west][inner sep=0.75pt]  [font=\normalsize]  {$\hat{\mathbf{p}}$};
\draw (214,80) node [anchor=north west][inner sep=0.75pt]  [font=\normalsize]  {$\hat{\mathbf{p}}_{\textrm o}$};

\draw (168,139) node [anchor=north west][inner sep=0.75pt]  [font=\normalsize] [align=left] {transfor\\mation};
\draw (91,135)node [anchor=north west][inner sep=0.75pt]  [font=\normalsize] [align=left] {decoder\\network\\\autoref{fig:nat}b};

\draw (233,135) node [anchor=north west][inner sep=0.75pt]  [font=\normalsize] [align=left] {encoder\\ network\\\autoref{fig:nat}a};

\draw (307,129) node [anchor=north west][inner sep=0.75pt]  [font=\normalsize] [align=left] {KPIs \\predictior \\ \autoref{fig:nat}c};

\draw (190,245) node [anchor=north west][inner sep=0.75pt]  [font=\normalsize] [align=left] {MOO (NSGA-\rom{2})};

\end{tikzpicture}}

%% file: fig/DNN_optimization_workflow.tex
\tikzset{every picture/.style={line width=0.75pt}} 
\resizebox{\linewidth}{!}{

\begin{tikzpicture}[x=0.75pt,y=0.75pt,yscale=-1,xscale=1]

\draw  [fill={rgb, 255:red, 155; green, 155; blue, 155 }  ,fill opacity=1 ] (285.6,232.22) -- (214.5,232.22) -- (167.1,219.05) -- (214.5,205.88) -- (285.6,205.88) -- cycle ;
\draw  [fill={rgb, 255:red, 126; green, 211; blue, 33 }  ,fill opacity=1 ] (313,96.1) -- (327,96.1) -- (327,166.23) -- (313,166.23) -- cycle ;
\draw    (320.8,166) -- (320.8,204) -- (320.8,220) -- (291.67,220.23) ;
\draw [shift={(288.67,220.25)}, rotate = 359.55] [fill={rgb, 255:red, 0; green, 0; blue, 0 }  ][line width=0.08]  [draw opacity=0] (5.36,-2.57) -- (0,0) -- (5.36,2.57) -- cycle    ;
\draw    (167.1,219.05) -- (97.1,219.05) -- (96.23,175.25) ;
\draw [shift={(96.17,172.25)}, rotate = 88.86] [fill={rgb, 255:red, 0; green, 0; blue, 0 }  ][line width=0.08]  [draw opacity=0] (5.36,-2.57) -- (0,0) -- (5.36,2.57) -- cycle    ;
\draw   (160.73,107.95) .. controls (160.73,99.17) and (167.85,92.05) .. (176.63,92.05) -- (264.63,92.05) .. controls (273.41,92.05) and (280.53,99.17) .. (280.53,107.95) -- (280.53,155.65) .. controls (280.53,164.43) and (273.41,171.55) .. (264.63,171.55) -- (176.63,171.55) .. controls (167.85,171.55) and (160.73,164.43) .. (160.73,155.65) -- cycle ;
\draw   (61.5,91.55) -- (131.5,91.55) -- (131.5,172.8) -- (61.5,172.8) -- cycle ;
\draw    (131.92,130.76) -- (157.4,130.77) ;
\draw [shift={(160.4,130.77)}, rotate = 180.01] [fill={rgb, 255:red, 0; green, 0; blue, 0 }  ][line width=0.08]  [draw opacity=0] (5.36,-2.57) -- (0,0) -- (5.36,2.57) -- cycle    ;
\draw    (280.4,130.43) -- (311,130.95) ;
\draw [shift={(314,131)}, rotate = 180.97] [fill={rgb, 255:red, 0; green, 0; blue, 0 }  ][line width=0.08]  [draw opacity=0] (5.36,-2.57) -- (0,0) -- (5.36,2.57) -- cycle    ;
\draw (224.4,136.6) node  {\includegraphics[width=73.5pt,height=49pt]{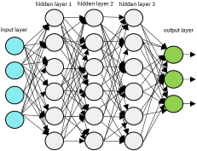}};

\draw (247,220) node  [font=\small] [align=left] {\begin{minipage}[lt]{77.97pt}\setlength\topsep{0pt}
MOO (NSGA-\rom 2)
\end{minipage}};
\draw (316.27,79) node [anchor=north west][inner sep=0.75pt]  [font=\small]  {$\hat{\mathbf{\kpi}}$};
\draw (95.77,130) node  [font=\scriptsize] [align=left] {\begin{minipage}[lt]{47.74pt}\setlength\topsep{0pt}
\begin{center}
Scalar input parameters of single machine class (ASM or PMSM)
\end{center}

\end{minipage}};
\draw (232,98.64) node  [font=\scriptsize] [align=left] {\begin{minipage}[lt]{72.85pt}\setlength\topsep{0pt}
deep neural network
\end{minipage}};

\end{tikzpicture}}

%% file: fig/network_architecture.tex
\tikzstyle{block_input_layers} = [rectangle, draw, text width=7em, text centered, rounded corners, minimum height=1em,font=\small ]
\tikzstyle{block_output_layers} = [rectangle, draw, fill = brown!30, text width=2.85em, text centered, minimum height=1em,font=\small ]
\tikzstyle{block_output_layers2} = [rectangle, draw, fill = green!30, 	text width=7em, text centered,  minimum height=1em,font=\small ]
\tikzstyle{block_intermediate_layers} = [rectangle, draw,     fill = blue!30, 	text width=7em, text centered, minimum height=1em,font=\small ]
\tikzstyle{dotted_block} = [draw=black!, line width=0.5pt, dash pattern=on 1pt off 4pt on 6pt off 4pt,inner ysep=0.8mm,inner xsep=0.8mm, rectangle, rounded corners]
\tikzstyle{line} = [draw, -latex']
\resizebox{\linewidth}{!}{
\begin{tikzpicture}[node distance = 0.70em and 0.7em, auto]

	\node [block_input_layers] (E0) {input vector($\mathbf{p}$): $52 \times 1$};
	\node [block_intermediate_layers, below =of E0] (E1) {$52\times5$ Conv1D, stride 1 + tanh};
	\node [block_intermediate_layers, below =of E1] (E2) {$52\times10$ Conv1D, stride 1 + tanh};
	\node [block_intermediate_layers, below =of E2] (E3) {$52\times10$ Conv1D, stride 1 + tanh};
	\node [block_intermediate_layers, below =of E3] (E4) {$52\times20$ Conv1D, stride 1 + tanh};
    \node [block_intermediate_layers, below =of E4] (E5) {1040 flatten layer };
    \node [block_intermediate_layers, below =of E5] (E6) {800 \ Dense, tanh};
    \node[block_output_layers, below right= 0.7em and -1.25cm of E6](out1){$\bm{\upsilon}$: 34 Dense};

    \node[block_output_layers, left =  of out1](out2){$\bm{\sigma}$: 34 Dense};
    
	\path [line] (E0) -- (E1); 
	\path [line] (E1) -- (E2); 
	\path [line] (E2) -- (E3); 
	\path [line] (E3) -- (E4); 
	\path [line] (E4) -- (E5); 
	\path [line] (E5) -- (E6); 

	\node [block_input_layers, right =of E0] (D0) {latent vector ($\mathbf{z}$): $34 \times 1$};
	\node [block_intermediate_layers, below =of D0] (D1) {800 \ Dense, tanh};
	\node [block_intermediate_layers, below =of D1] (D2) {1040 \ Dense, tanh};
	\node [block_intermediate_layers, below =of D2] (D3) {$52\times20$ Cov1DT, stride 1 + tanh};
    \node [block_intermediate_layers, below =of D3] (D4) {$52\times10$ Cov1DT, stride 1 + tanh};
    \node [block_intermediate_layers, below =of D4] (D5) {$52\times10$ Cov1DT, stride 1 + tanh};
    \node [block_intermediate_layers, below =of D5] (D6) {$52\times5$ Cov1DT, stride 1 + tanh};
    \node [block_intermediate_layers, below =of D6] (D7) {$32\times1$ Cov1DT, stride 1 + linear};
    \node [block_output_layers2, below =of D7] (D8) {pred.parameters ($\hat{\mathbf{p}}$): 52};
    \node [block_output_layers2, left =of D8] (E8) {latent vector ($\mathbf{z}$): $34 \times 1$};
   
    \node(v0) at ([xshift=0.73cm, yshift=-0.4em]E8.north){};
    \draw [line] (out1.south) -- (v0);
     \node(v1) at ([xshift=-0.76cm, yshift=-0.4em]E8.north){};
    \draw [line] (out2.south) -- (v1);
    
    \node(v2) at ([xshift=0.76cm, yshift=-0.88em]E6.south){};
    \draw [line] (E6) -- (v2);
     \node(v3) at ([xshift=-0.76cm, yshift=-0.88em]E6.south){};
    \draw [line] (E6) -- (v3);

    \path [line] (D0) -- (D1); 
	\path [line] (D1) -- (D2); 
	\path [line] (D2) -- (D3); 
	\path [line] (D3) -- (D4); 
	\path [line] (D4) -- (D5); 
	\path [line] (D5) -- (D6); 
	\path [line] (D6) -- (D7); 
	\path [line] (D7) -- (D8); 
    \node [dotted_block, fit = (D1) (D7)] (de) {};
    \node [dotted_block, fit = (E1) (out1) (out2)] (ee) {};

    \node [block_input_layers, right =of D0] (K0) {latent vector ($\mathbf{z}$): $34 \times 1$};
    \node [block_intermediate_layers, below =of K0] (K1) {448 \ Dense, softplus};
	\node [block_intermediate_layers, below =of K1] (K2) {250 \ Dense, softplus};
	\node [block_intermediate_layers, below =of K2] (K3) {224 \ Dense, softplus};
	\node [block_intermediate_layers, below =of K3] (K4) {224 \ Dense, softplus};
    \node [block_intermediate_layers, below =of K4] (K5) {198 \ Dense, softplus};
    \node [block_intermediate_layers, below =of K5] (K6) {50 \ Dense, softplus};
    \node [block_output_layers2, below =of K6] (K7) {output KPIs ($\hat{\mathbf{k}}$): 3};
    
    \path [line] (K0) -- (K1); 
	\path [line] (K1) -- (K2); 
	\path [line] (K2) -- (K3); 
	\path [line] (K3) -- (K4); 
	\path [line] (K4) -- (K5); 
	\path [line] (K5) -- (K6); 
	\path [line] (K6) -- (K7); 
    \node [dotted_block, fit = (K1) (K6)] (ke) {};

    \node(v4) at ([xshift=-0.0cm, yshift=-0.7em]E8.south) [font=\normalsize]{a) Encoder};
    \node(v5) at ([xshift=-0.0cm, yshift=-0.7em]D8.south) [font=\normalsize]{b) Decoder};
    \node(v6) at ([xshift=-0.0cm, yshift=-0.7em]K7.south)[font=\normalsize]{c) KPIs Predictor};

\end{tikzpicture}}

%% file: fig/validation_loss.tex

\begin{tikzpicture}
\tikzstyle{every node}=[font=\footnotesize]

\definecolor{color0}{rgb}{0.12156862745098,0.466666666666667,0.705882352941177}
\definecolor{color1}{rgb}{1,0.498039215686275,0.0549019607843137}

\begin{axis}[
legend cell align={left},
ymode=log,
width=0.85\linewidth,
height=4cm,
legend style={fill opacity=0.8, draw opacity=1, text opacity=1, draw=white!80!black},
tick align=outside,
tick pos=left,
title={Evaluation},
x grid style={white!69.0196078431373!black},
xlabel={No of Epochs},
xmin=-14.95, xmax=313.95,
xtick style={color=black},
xtick={0,50,100,150,200,250,300,350},
xticklabels={0,50,100,150,200,250,300,350},
y grid style={white!69.0196078431373!black},
ylabel={Total\_loss},
ymin=-0.0438048920594156, ymax=1.0232610876672,
]
\addplot [semithick, color0]
table {%
0 0.583875715732574
1 0.117019847035408
2 0.0907157510519028
3 0.060718446969986
4 0.0464825853705406
5 0.0396114252507687
6 0.0351591557264328
7 0.0317887328565121
8 0.0310478899627924
9 0.0305807199329138
10 0.0299874916672707
11 0.0296520330011845
12 0.0291059482842684
13 0.0285840053111315
14 0.0280528981238604
15 0.0274391379207373
16 0.0270325765013695
17 0.0261594709008932
18 0.0251022130250931
19 0.0243167337030172
20 0.0235626399517059
21 0.0231480523943901
22 0.0230518896132708
23 0.0226070303469896
24 0.0223005246371031
25 0.022112974897027
26 0.0219280179589987
27 0.0216444209218025
28 0.021456541493535
29 0.0214237328618765
30 0.0212910529226065
31 0.0210421737283468
32 0.0209961552172899
33 0.0207818392664194
34 0.0206912662833929
35 0.02068299241364
36 0.0204733684659004
37 0.0203479286283255
38 0.0201940126717091
39 0.0201847832649946
40 0.0201634149998426
41 0.0200126711279154
42 0.0197948459535837
43 0.0197414290159941
44 0.0197442602366209
45 0.0195935349911451
46 0.0194899775087833
47 0.0194560885429382
48 0.0193493831902742
49 0.0192463900893927
50 0.0191940609365702
51 0.0192422848194838
52 0.0192254167050123
53 0.0191358085721731
54 0.0190240181982517
55 0.0189444124698639
56 0.0188317168504
57 0.0188550148159266
58 0.0188009571284056
59 0.0187152251601219
60 0.0192947573959827
61 0.0186199061572552
62 0.0186676438897848
63 0.0185324847698212
64 0.018548496067524
65 0.0185196660459042
66 0.0183868072926998
67 0.0184171907603741
68 0.0183345768600702
69 0.0183337200433016
70 0.0183128956705332
71 0.0182751473039389
72 0.0184287521988153
73 0.0181717220693827
74 0.0181212704628706
75 0.01819814927876
76 0.0180619526654482
77 0.0180758852511644
78 0.0179752185940742
79 0.0178969781845808
80 0.0179198700934649
81 0.0180812831968069
82 0.0178699605166912
83 0.0178294647485018
84 0.0179181899875402
85 0.0177941620349884
86 0.0177047662436962
87 0.0178026184439659
88 0.0177515614777803
89 0.0183888655155897
90 0.0177288632839918
91 0.0176369827240705
92 0.0177117865532637
93 0.0176775231957436
94 0.017769567668438
95 0.0175889953970909
96 0.0175995156168938
97 0.0176580157130957
98 0.0175698772072792
99 0.0175883136689663
100 0.0174788404256105
101 0.0174927208572626
102 0.0174786113202572
103 0.0175075586885214
104 0.0174280479550362
105 0.0175144486129284
106 0.0173987615853548
107 0.0174211487174034
108 0.0174388326704502
109 0.0174586325883865
110 0.0173748731613159
111 0.0173453595489264
112 0.0173296518623829
113 0.017401996999979
114 0.0172672308981419
115 0.0172852352261543
116 0.0172529332339764
117 0.0172497723251581
118 0.017241733148694
119 0.0176527667790651
120 0.017222385853529
121 0.0171635858714581
122 0.0171577017754316
123 0.018573509529233
124 0.0171906836330891
125 0.0171541050076485
126 0.0171830896288157
127 0.0171730369329453
128 0.0171475000679493
129 0.0175454318523407
130 0.0175664853304625
131 0.0171584449708462
132 0.0170784648507833
133 0.0170958694070578
134 0.0175862647593021
135 0.0170422997325659
136 0.0171192940324545
137 0.0169841889292002
138 0.0173760522156954
139 0.0169994439929724
140 0.0172467865049839
141 0.017014667391777
142 0.0169682539999485
143 0.0169797912240028
144 0.0170037318021059
145 0.0168882012367249
146 0.0169228538870811
147 0.0169142335653305
148 0.0173843838274479
149 0.0178654976189137
150 0.0169561579823494
151 0.0169124398380518
152 0.01693045347929
153 0.0173458270728588
154 0.0175401102751493
155 0.0170227121561766
156 0.0168222859501839
157 0.0168677903711796
158 0.0169319175183773
159 0.0168279260396957
160 0.0167889930307865
161 0.0167936179786921
162 0.0168179459869862
163 0.0167138781398535
164 0.016830075532198
165 0.0167279783636332
166 0.0167035013437271
167 0.016786452382803
168 0.0167571976780891
169 0.0166953559964895
170 0.0167403388768435
171 0.0167189389467239
172 0.0166694279760122
173 0.0167193226516247
174 0.0166817102581263
175 0.0166593994945288
176 0.0166843049228191
177 0.0166174788028002
178 0.0166217945516109
179 0.0165949165821075
180 0.0166221130639315
181 0.0165928304195404
182 0.0165600199252367
183 0.0165794752538204
184 0.0166920460760593
185 0.0165496822446585
186 0.0167517010122538
187 0.0165320076048374
188 0.0166169982403517
189 0.0165245030075312
190 0.0166586935520172
191 0.0165091827511787
192 0.016564754769206
193 0.0165143292397261
194 0.0165646448731422
195 0.0164697859436274
196 0.0165047813206911
197 0.0166472177952528
198 0.016903443261981
199 0.0164880026131868
200 0.0165094193071127
201 0.0164529997855425
202 0.0164340268820524
203 0.0164430756121874
204 0.0164492949843407
205 0.0164598356932402
206 0.0163794551044703
207 0.0164572019129992
208 0.0164052210748196
209 0.0163808837532997
210 0.0163671523332596
211 0.0163316503167152
212 0.016414163634181
213 0.016902131959796
214 0.0168491415679455
215 0.0164058730006218
216 0.0163584481924772
217 0.0163098108023405
218 0.0164342280477285
219 0.0167331621050835
220 0.0170162729918957
221 0.016368554905057
222 0.0163537170737982
223 0.0163141898810863
224 0.0163163300603628
225 0.0163332317024469
226 0.0163614116609097
227 0.0163172725588083
228 0.0162711255252361
229 0.0163177940994501
230 0.0162925254553556
231 0.0162956286221743
232 0.016251266002655
233 0.0162582695484161
234 0.0162086915224791
235 0.0161949992179871
236 0.0162374563515186
237 0.0162284076213837
238 0.0161786153912544
239 0.0161934532225132
240 0.0163164474070072
241 0.0161687973886728
242 0.0162926074117422
243 0.0168818682432175
244 0.016192864626646
245 0.0160882603377104
246 0.0164783634245396
247 0.0161220878362656
248 0.0161285884678364
249 0.0162985846400261
250 0.0161813609302044
251 0.0161262154579163
252 0.0161441564559937
253 0.0161070320755243
254 0.0161285698413849
255 0.0160907302051783
256 0.0166086051613092
257 0.0163823682814837
258 0.0161565467715263
259 0.0160403214395046
260 0.0161443650722504
261 0.0165818240493536
262 0.0161086954176426
263 0.0160521771758795
264 0.0160791985690594
265 0.0160691719502211
266 0.0165912080556154
267 0.0161635149270296
268 0.0160840768367052
269 0.0160325225442648
270 0.0160429794341326
271 0.0166491586714983
272 0.0165309049189091
273 0.0159888379275799
274 0.0160373952239752
275 0.0159921254962683
276 0.0165437757968903
277 0.0160031300038099
278 0.0159625131636858
279 0.0160385500639677
280 0.0159640479832888
281 0.0163211692124605
282 0.0160227306187153
283 0.0159461833536625
284 0.0159787945449352
285 0.0159575492143631
286 0.0160611961036921
287 0.0159434732049704
288 0.0160500407218933
289 0.0163863562047482
290 0.0162704735994339
291 0.0159463714808226
292 0.0159083884209394
293 0.0159246399998665
294 0.0159927550703287
295 0.0159133449196815
296 0.015886127948761
297 0.016118548810482
298 0.0159717556089163
299 0.0159104447811842
};
\addlegendentry{training_loss}
\addplot [semithick, color1]
table {%
0 0.130238503217697
1 0.14874267578125
2 0.0986223518848419
3 0.0524865426123142
4 0.0465968698263168
5 0.0436058007180691
6 0.0346482321619987
7 0.0329628698527813
8 0.0304008033126593
9 0.0315942093729973
10 0.0313323214650154
11 0.0296173244714737
12 0.0283929090946913
13 0.0290942750871181
14 0.0286587122827768
15 0.0292212069034576
16 0.0269614066928625
17 0.0262860730290413
18 0.0248294658958912
19 0.0247083902359009
20 0.0229177363216877
21 0.0266097988933325
22 0.0227441657334566
23 0.0224805809557438
24 0.02283008210361
25 0.0216702967882156
26 0.0225967150181532
27 0.0215018168091774
28 0.0219984389841557
29 0.0209677573293447
30 0.021169513463974
31 0.0221631936728954
32 0.0211592335253954
33 0.0212749186903238
34 0.0210429411381483
35 0.0203074514865875
36 0.0212992336601019
37 0.0201847814023495
38 0.0211272295564413
39 0.0200957972556353
40 0.0201656632125378
41 0.0200802683830261
42 0.0196700431406498
43 0.0228961929678917
44 0.0206177644431591
45 0.0198279991745949
46 0.0192631781101227
47 0.021164795383811
48 0.0200436394661665
49 0.0195149816572666
50 0.019564313814044
51 0.0197855401784182
52 0.0195890590548515
53 0.0191784743219614
54 0.0188834108412266
55 0.0197219792753458
56 0.0196238365024328
57 0.0192833505570889
58 0.020010931417346
59 0.0193286743015051
60 0.0185318030416965
61 0.0196930766105652
62 0.0183847770094872
63 0.0192933324724436
64 0.0186283830553293
65 0.0195050723850727
66 0.0201946180313826
67 0.0183997452259064
68 0.0185934212058783
69 0.0190697591751814
70 0.0182886105030775
71 0.0213035959750414
72 0.01790533028543
73 0.019529664888978
74 0.0184779670089483
75 0.0191816948354244
76 0.0181066691875458
77 0.0183984022587538
78 0.0197294373065233
79 0.0178695060312748
80 0.0187134146690369
81 0.0179006308317184
82 0.0179749727249146
83 0.0182263571768999
84 0.0183379482477903
85 0.0185991022735834
86 0.0178415961563587
87 0.0178793780505657
88 0.0180915128439665
89 0.0178365334868431
90 0.0184215400367975
91 0.0180733483284712
92 0.0173988379538059
93 0.0178239941596985
94 0.018940132111311
95 0.0175970885902643
96 0.0174170006066561
97 0.0179596152156591
98 0.0177782047539949
99 0.0174569729715586
100 0.0179098751395941
101 0.0175809022039175
102 0.0175277814269066
103 0.0173070579767227
104 0.0180613957345486
105 0.0174331441521645
106 0.0186072699725628
107 0.0180383827537298
108 0.0175301060080528
109 0.0178381912410259
110 0.0173471830785275
111 0.0177189279347658
112 0.0181711446493864
113 0.0176296234130859
114 0.0174034778028727
115 0.0179911740124226
116 0.0173652451485395
117 0.0170392394065857
118 0.0176227446645498
119 0.0173986405134201
120 0.0177684668451548
121 0.0173727832734585
122 0.0174394492059946
123 0.0175573751330376
124 0.0173138584941626
125 0.0172677580267191
126 0.0171677302569151
127 0.0177619755268097
128 0.0186030399054289
129 0.0176401380449533
130 0.0172527860850096
131 0.0172778330743313
132 0.0174216646701097
133 0.0181075874716043
134 0.0170572195202112
135 0.017592316493392
136 0.0175606086850166
137 0.0171301420778036
138 0.0173904560506344
139 0.0171277672052383
140 0.0173305366188288
141 0.017091691493988
142 0.0169184356927872
143 0.0172539874911308
144 0.0172167420387268
145 0.0171720925718546
146 0.0181436203420162
147 0.0174856279045343
148 0.0175599399954081
149 0.0170884970575571
150 0.0171380415558815
151 0.0170311685651541
152 0.0170685015618801
153 0.0168898869305849
154 0.0177132189273834
155 0.0179235078394413
156 0.0169793404638767
157 0.0178389176726341
158 0.0192898791283369
159 0.0171826668083668
160 0.0169233828783035
161 0.0171812158077955
162 0.0171663369983435
163 0.01698755659163
164 0.0168443061411381
165 0.0173346325755119
166 0.0169627368450165
167 0.0173994600772858
168 0.0168878100812435
169 0.017191244289279
170 0.0172419995069504
171 0.0174215883016586
172 0.0170193444937468
173 0.0170136541128159
174 0.016728151589632
175 0.0170009415596724
176 0.0166231878101826
177 0.0169155914336443
178 0.0169264879077673
179 0.0168730579316616
180 0.0167853627353907
181 0.0176171287894249
182 0.017304515466094
183 0.0166574846953154
184 0.0174920056015253
185 0.0168512295931578
186 0.0167020168155432
187 0.0167673025280237
188 0.0167100764811039
189 0.0166546553373337
190 0.0165183693170547
191 0.0166129507124424
192 0.0166738387197256
193 0.0169944632798433
194 0.0171023309230804
195 0.0169087518006563
196 0.0165381617844105
197 0.0169925205409527
198 0.0171176567673683
199 0.0164491981267929
200 0.0173912327736616
201 0.0170081593096256
202 0.0171089973300695
203 0.0166614409536123
204 0.0170757789164782
205 0.0166554916650057
206 0.0165557134896517
207 0.0168018452823162
208 0.0167980678379536
209 0.0168962813913822
210 0.0167079158127308
211 0.0167231597006321
212 0.0165380407124758
213 0.0165113527327776
214 0.0164966732263565
215 0.0166504625231028
216 0.0166798233985901
217 0.0194142907857895
218 0.0166301149874926
219 0.0168655905872583
220 0.0167616046965122
221 0.0167063418775797
222 0.0165298990905285
223 0.0166949480772018
224 0.016532065346837
225 0.0172198582440615
226 0.0167182721197605
227 0.0166984982788563
228 0.0168173555284739
229 0.0168584864586592
230 0.0166954044252634
231 0.0165302213281393
232 0.0168289374560118
233 0.0163938235491514
234 0.0165551658719778
235 0.0163749251514673
236 0.0166942365467548
237 0.0165570937097073
238 0.016868194565177
239 0.0165263973176479
240 0.0165083725005388
241 0.0169480219483376
242 0.0171029027551413
243 0.0169667154550552
244 0.0163226127624512
245 0.0164753161370754
246 0.0165680181235075
247 0.0165550746023655
248 0.0173350069671869
249 0.0164579953998327
250 0.0162284448742867
251 0.0163871329277754
252 0.0167283173650503
253 0.0167189575731754
254 0.0162907969206572
255 0.0162863526493311
256 0.0168801359832287
257 0.0164521764963865
258 0.0163681693375111
259 0.0164102707058191
260 0.0161695946007967
261 0.0163483712822199
262 0.0163098331540823
263 0.0164862908422947
264 0.0162040609866381
265 0.0166207905858755
266 0.0164335649460554
267 0.0165891852229834
268 0.0161027554422617
269 0.0168734211474657
270 0.0165395233780146
271 0.0163720156997442
272 0.0161871071904898
273 0.0164161566644907
274 0.0171101056039333
275 0.0160425137728453
276 0.0162145774811506
277 0.0162757504731417
278 0.0162851195782423
279 0.0165184084326029
280 0.0165575463324785
281 0.0162840858101845
282 0.016720762476325
283 0.01680913195014
284 0.0163507629185915
285 0.0161098185926676
286 0.0159487053751945
287 0.0166500676423311
288 0.0163255780935287
289 0.016245486214757
290 0.0163065828382969
291 0.0162361655384302
292 0.0160638559609652
293 0.0164982974529266
294 0.0162686705589294
295 0.0161739978939295
296 0.0162315517663956
297 0.0164937376976013
298 0.0163363106548786
299 0.0166212152689695
};
\addlegendentry{validation_loss}
\end{axis}

\end{tikzpicture}

%% file: fig/vae_mlp_cmp.tex
\resizebox{\linewidth}{!}{
\begin{tikzpicture}

\definecolor{color0}{rgb}{0.12156862745098,0.466666666666667,0.705882352941177}
\definecolor{color1}{rgb}{1,0.498039215686275,0.0549019607843137}
\tikzstyle{every node}=[font=\normalsize]
\pgfplotsset{every axis title/.append style={at={(0.5,0.93)}}}

\begin{groupplot}[group style={group size=2 by 1,
                 horizontal sep = 2 cm, 
            vertical sep = 0.8cm}, 
        width = 4.5 cm, 
        height = 4.5 cm,]
\nextgroupplot[
legend cell align={left},
legend style={fill opacity=0.8, draw opacity=1, text opacity=1, at={(1.52,1)}, draw=white!80!black},
tick align=outside,
tick pos=left,
title={ASM},
x grid style={white!69.0196078431373!black},
xlabel={KPIs},
xmin=-0.21, xmax=2.21,
xtick style={color=black},
xtick={0,1,2},
xticklabels={\(\displaystyle k_{1}\),\(\displaystyle k_{2}\),\(\displaystyle k_{3}\)},
y grid style={white!69.0196078431373!black},
ylabel={Mean absolute error},
ymin=0, ymax=6.678,
ytick style={color=black}
]
\draw[draw=none,fill=color0] (axis cs:-0.1,0) rectangle (axis cs:0.1,0.26);
\addlegendimage{ybar,ybar legend,draw=none,fill=color0}
\addlegendentry{DNN}

\draw[draw=none,fill=color0] (axis cs:0.9,0) rectangle (axis cs:1.1,2.28);
\draw[draw=none,fill=color0] (axis cs:1.9,0) rectangle (axis cs:2.1,6.36);
\draw[draw=none,fill=color1] (axis cs:-0.1,0) rectangle (axis cs:0.1,0.49);
\addlegendimage{ybar,ybar legend,draw=none,fill=color1}
\addlegendentry{VAE}

\draw[draw=none,fill=color1] (axis cs:0.9,0) rectangle (axis cs:1.1,2.12);
\draw[draw=none,fill=color1] (axis cs:1.9,0) rectangle (axis cs:2.1,5.76);

\nextgroupplot[
tick align=outside,
tick pos=left,
title={PMSM},
x grid style={white!69.0196078431373!black},
xlabel={KPIs},
xmin=-0.21, xmax=2.21,
xtick style={color=black},
xtick={0,1,2},
xticklabels={\(\displaystyle k_{1}\),\(\displaystyle k_{2}\),\(\displaystyle k_{3}\)},
y grid style={white!69.0196078431373!black},
ymin=0, ymax=2.856,
ytick style={color=black}
]
\draw[draw=none,fill=color0] (axis cs:-0.1,0) rectangle (axis cs:0.1,0.42);
\draw[draw=none,fill=color0] (axis cs:0.9,0) rectangle (axis cs:1.1,2.09);
\draw[draw=none,fill=color0] (axis cs:1.9,0) rectangle (axis cs:2.1,2.72);
\draw[draw=none,fill=color1] (axis cs:-0.1,0) rectangle (axis cs:0.1,0.37);
\draw[draw=none,fill=color1] (axis cs:0.9,0) rectangle (axis cs:1.1,1.68);
\draw[draw=none,fill=color1] (axis cs:1.9,0) rectangle (axis cs:2.1,2.2);
\end{groupplot}

\end{tikzpicture}}

%% file: fig/Pareto_design_visualization.tex
\tikzset{every picture/.style={line width=0.75pt}} 
\resizebox{\linewidth}{!}{

\begin{tikzpicture}[x=0.75pt,y=0.75pt,yscale=-1,xscale=1]

\draw (105.26,152) node  {\includegraphics[width=140pt,height=130pt]{fig/Pareto_VAE_PSM1.pdf}};
\draw (280,170) node  {\includegraphics[width=100pt,height=100pt]{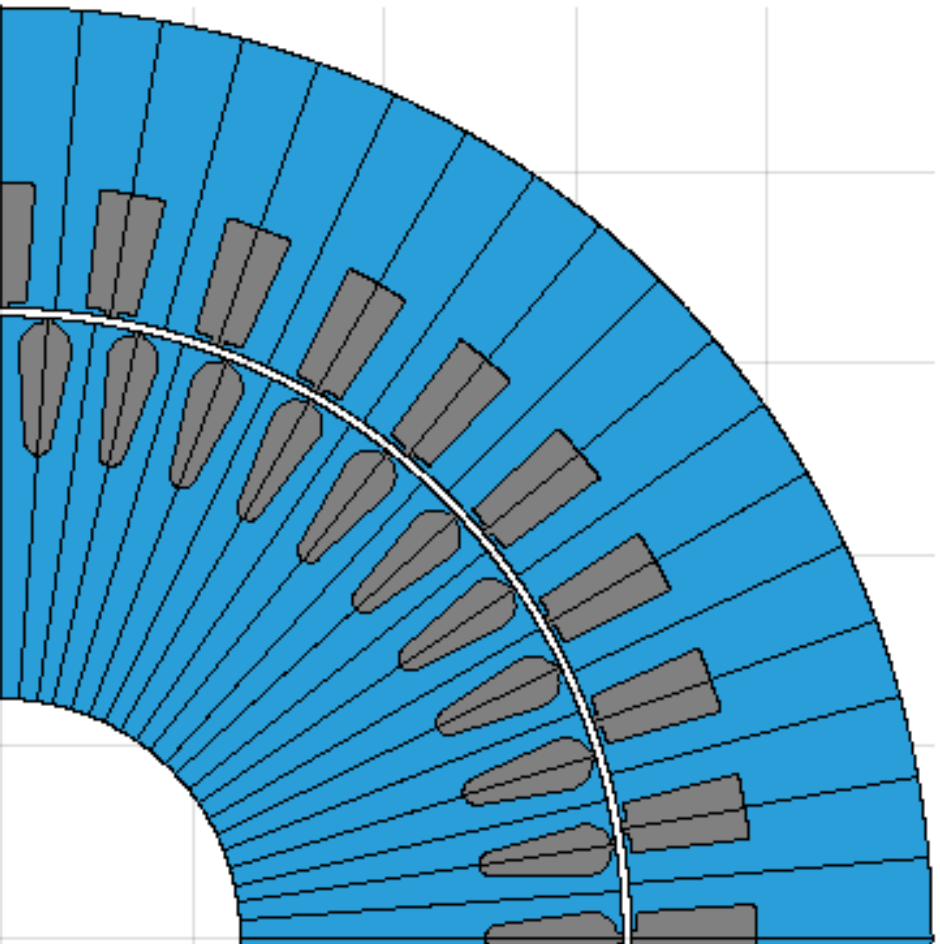}};
\draw (440,155) node  {\includegraphics[width=140pt,height=125pt]{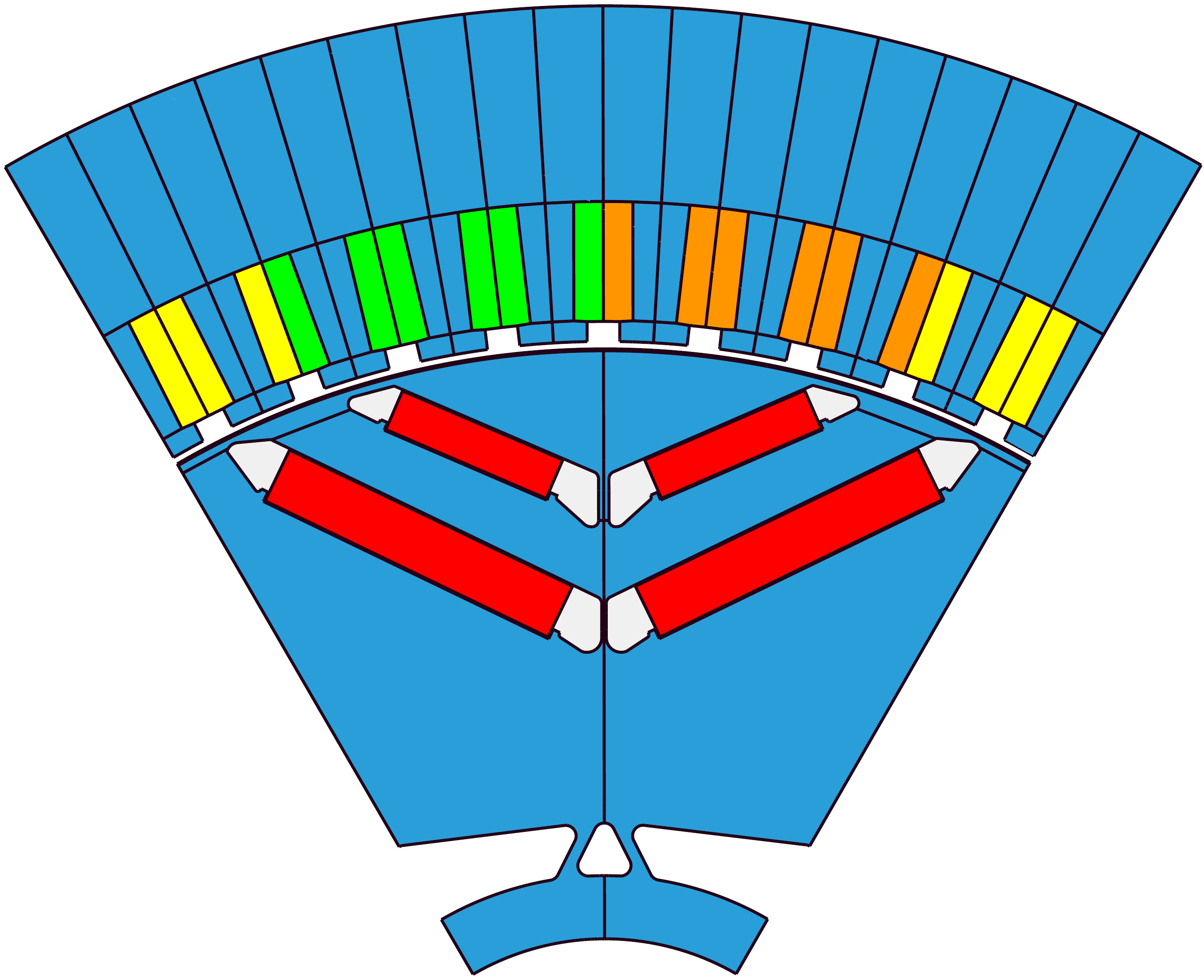}};
\draw (618,162) node  {\includegraphics[width=112.29pt,height=112.29pt]{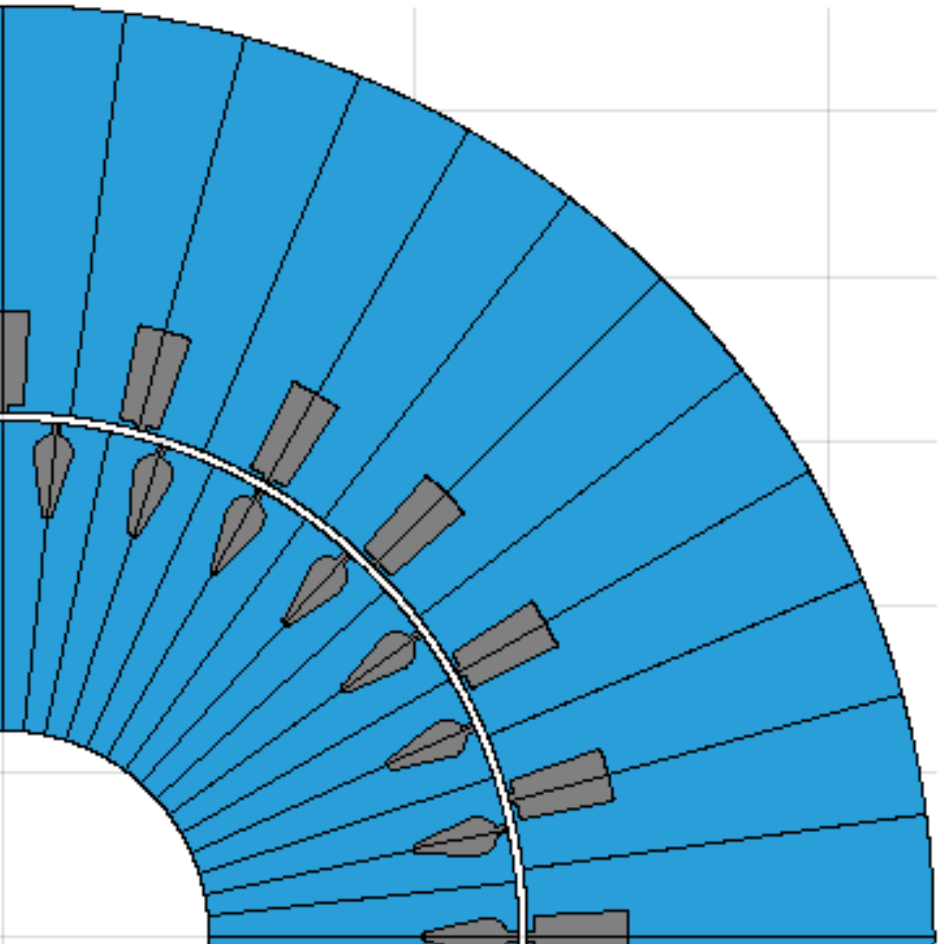}};

\draw (78,242) node [anchor=north west][inner sep=0.75pt]   [align=left] {Design A};
\draw (258,242) node [anchor=north west][inner sep=0.75pt]   [align=left] {Design B};
\draw (414,242) node [anchor=north west][inner sep=0.75pt]   [align=left] {Design C};
\draw (582,242) node [anchor=north west][inner sep=0.75pt]   [align=left] {Design D};

\end{tikzpicture}}

%% file: fig/Pareto_front_all_combined.tex
\tikzset{every picture/.style={line width=0.75pt}} 

\begin{tikzpicture}[x=0.75pt,y=0.75pt,yscale=-1,xscale=1]
\draw (272.31,140.82) node  {\includegraphics[width=332.53pt,height=200.51pt]{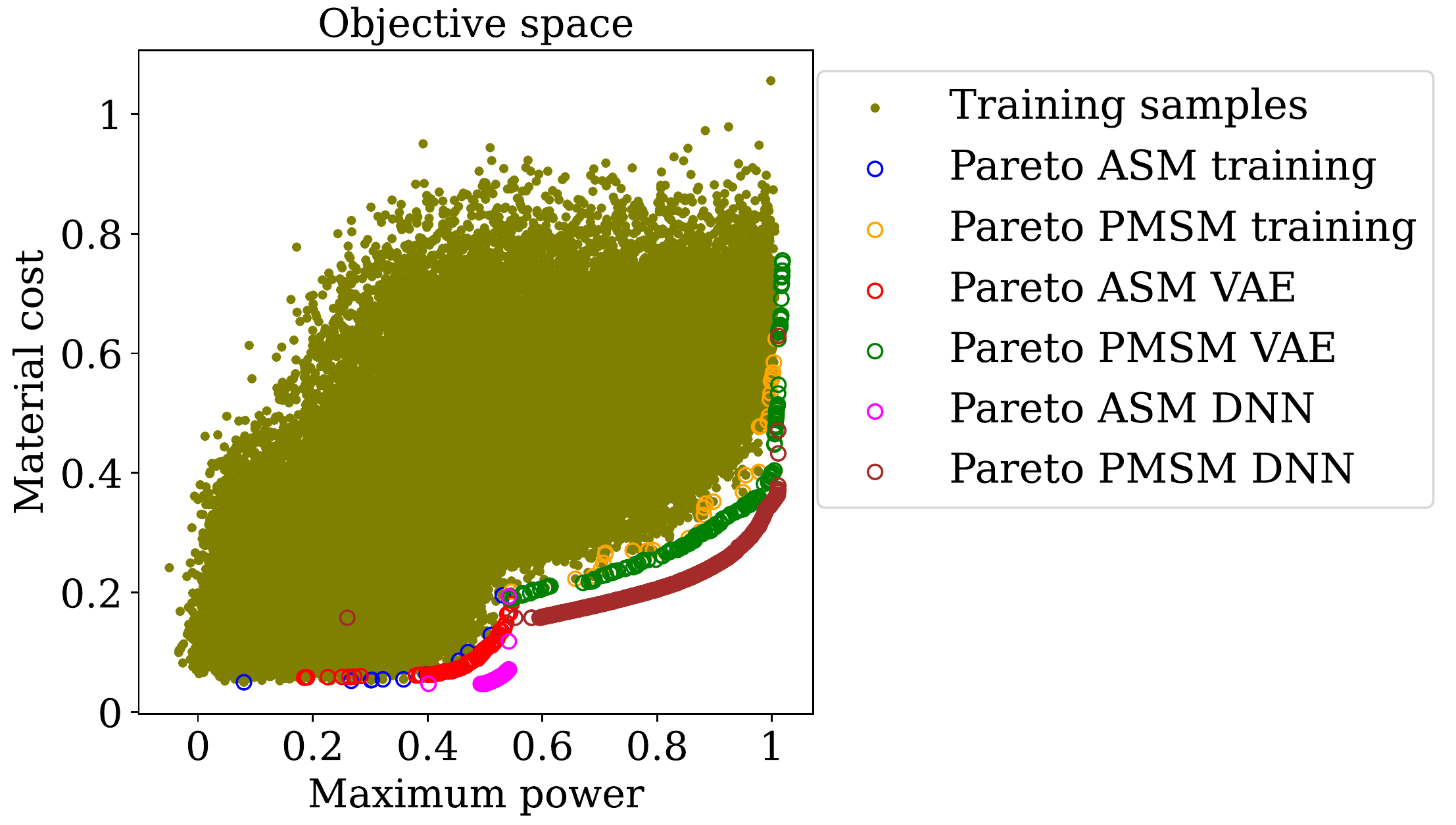}};
\draw    (125.07,161.09) -- (204.48,199.33) ;
\draw [shift={(207.18,200.64)}, rotate = 205.71] [fill={rgb, 255:red, 0; green, 0; blue, 0 }  ][line width=0.08]  [draw opacity=0] (5.36,-2.57) -- (0,0) -- (5.36,2.57) -- cycle    ;
\draw    (219.9,64.15) -- (286.19,89.27) ;
\draw [shift={(289,90.33)}, rotate = 200.75] [fill={rgb, 255:red, 0; green, 0; blue, 0 }  ][line width=0.08]  [draw opacity=0] (5.36,-2.57) -- (0,0) -- (5.36,2.57) -- cycle    ;
\draw    (283.9,222.25) -- (209.66,203.85) ;
\draw [shift={(206.75,203.13)}, rotate = 13.92] [fill={rgb, 255:red, 0; green, 0; blue, 0 }  ][line width=0.08]  [draw opacity=0] (5.36,-2.57) -- (0,0) -- (5.36,2.57) -- cycle    ;
\draw    (328.73,184.42) -- (292.23,120.92) ;
\draw [shift={(290.73,118.32)}, rotate = 60.11] [fill={rgb, 255:red, 0; green, 0; blue, 0 }  ][line width=0.08]  [draw opacity=0] (5.36,-2.57) -- (0,0) -- (5.36,2.57) -- cycle    ;

\draw (129,154) node  [font=\small] [align=left] {\begin{minipage}[lt]{41.48pt}\setlength\topsep{0pt}
Design B
\end{minipage}};
\draw (192.43,58.8) node  [font=\small] [align=left] {\begin{minipage}[lt]{41.48pt}\setlength\topsep{0pt}
Design A
\end{minipage}};
\draw (343.9,189) node  [font=\small] [align=left] {\begin{minipage}[lt]{41.48pt}\setlength\topsep{0pt}
Design C
\end{minipage}};
\draw (311.86,222.87) node  [font=\small] [align=left] {\begin{minipage}[lt]{41.48pt}\setlength\topsep{0pt}
Design D
\end{minipage}};

\end{tikzpicture}

%% file: main_ref.bib
@ARTICLE{8661767,
		author={Khan, Arbaaz and Ghorbanian, Vahid and Lowther, David},
	    journal={IEEE Transactions on Magnetics}, 
        title={Deep Learning for Magnetic Field Estimation}, 
        year={2019},
        volume={55},
        number={6},
        pages={1-4},
        doi={10.1109/TMAG.2019.2899304}
		}

@INPROCEEDINGS{9940692,
               author={Cesay, Saikou and Teng, Paul and Wang, Ruoli and Yue, Haupeng and Khan, Arbaaz and Lowther, David},
               booktitle={2022 IEEE 20th Biennial Conference on Electromagnetic Field Computation (CEFC)}, 
               title={Generalizable DNN based multi-material Hysteresis Modelling}, 
               year={2022},
               volume={},
               number={},
               pages={1-2},
               doi={10.1109/CEFC55061.2022.9940692}
			   }

@INPROCEEDINGS{9291033,
               author={Quabeck, Stefan and Shangguan, Wenbo and Scharfenstein, Daniel and De Doncker, Rik W.},
               booktitle={2020 23rd International Conference on Electrical Machines and Systems (ICEMS)}, 
               title={Detection of Broken Rotor Bars in Induction Machines using Machine Learning Methods}, 
               year={2020},
               volume={},
               number={},
               pages={620-625},
               doi={10.23919/ICEMS50442.2020.9291033}
               }

@ARTICLE{9829204,
	     author={Pietrzak, Przemyslaw and Wolkiewicz, Marcin and Orlowska-Kowalska, Teresa},
         journal={IEEE Transactions on Industrial Electronics}, 
         title={PMSM Stator Winding Fault Detection and Classification Based on Bispectrum Analysis and Convolutional Neural Network}, 
         year={2023},
         volume={70},
         number={5},
         pages={5192-5202},
         doi={10.1109/TIE.2022.3189076}
		 }

@INPROCEEDINGS{9910710,
               author={Fatemimoghadam, Armita and Yan, Ye and Iyer, Lakshmi Varaha and Kar, Narayan C.},
               booktitle={2022 International Conference on Electrical Machines (ICEM)}, 
               title={Permanent Magnet Synchronous Motor Drive Using Deep-Neural-Network-Based Vector Control for Electric Vehicle Applications}, 
               year={2022},
               volume={},
               number={},
               pages={2358-2364},
               doi={10.1109/ICEM51905.2022.9910710}
			   }

@INPROCEEDINGS{8921886,
               author={Yan, Yu-Bai and Liang, Jia-Ning and Sun, Tian-Fu and Geng, Jian-Ping and Gang-Xie and Pan, Dong-Jia},
               booktitle={2019 22nd International Conference on Electrical Machines and Systems (ICEMS)}, 
               title={Torque Estimation and Control of PMSM Based on Deep Learning}, 
               year={2019},
               volume={},
               number={},
               pages={1-6},
               doi={10.1109/ICEMS.2019.8921886}
			   }

@INPROCEEDINGS{8056321,
               author={Jin, Liang and Wang, Fei and Yang, Qingxin},
               booktitle={2017 20th International Conference on Electrical Machines and Systems (ICEMS)}, 
               title={Performance analysis and optimization of permanent magnet synchronous motor based on deep learning}, 
               year={2017},
               volume={},
               number={},
               pages={1-5},
               doi={10.1109/ICEMS.2017.8056321}
			   }

@INPROCEEDINGS{9982881,
               author={Talukder, Akm Khaled Ahsan and Wang, Bingnan and Sakamoto, Yusuke},
               booktitle={2022 25th International Conference on Electrical Machines and Systems (ICEMS)}, 
               title={Electric Machine Two-dimensional Flux Map Prediction with Ensemble Learning}, 
               year={2022},
               volume={},
               number={},
               pages={1-4},
               doi={10.1109/ICEMS56177.2022.9982881}
			   }

@ARTICLE{9333549,  
        author = {Parekh, Vivek and Flore, Dominik and Schöps, Sebastian}, 
        journal = {IEEE Access},  
        title = {Deep Learning-Based Prediction of Key Performance Indicators for Electrical Machines}, 
        year = {2021}, 
        volume = {9}, 
        number = {},  
        pages = {21786-21797}, 
        doi = {10.1109/ACCESS.2021.3053856}
        }

@ARTICLE{9745548,
         author={Parekh, Vivek and Flore, Dominik and Schöps, Sebastian},
         journal={IEEE Transactions on Magnetics}, 
         title={Variational Autoencoder-Based Metamodeling for Multi-Objective Topology Optimization of Electrical Machines}, 
	     year={2022},
		 volume={58},
		 number={9},
		 pages={1-4},
		 doi={10.1109/TMAG.2022.3163972}}

@ARTICLE{MAL-056,
        author = {Diederik P. Kingma and Max Welling},
        journal = {Foundations and Trends® in Machine Learning},
        title = {An Introduction to Variational Autoencoders},
        year = {2019},
        volume = {12},
        number = {4},
        pages = {307-392},
        doi = {10.1561/2200000056},
        }

@ARTICLE{996017,
        author = {Deb, K. and Pratap, A. and Agarwal, S. and Meyarivan, T.},
        journal = {IEEE Transactions on Evolutionary Computation}, 
        title = {A fast and elitist multiobjective genetic algorithm: {NSGA}-{II}}, 
        year = {2002},
        volume = {6},
        number = {2},
        pages = {182-197},
        doi = {10.1109/4235.996017}
        }

@book{Salon_1995aa,
     author = "Salon, Sheppard J.",
     publisher = "Kluwer",
     year = "1995",
     title = "Finite Element Analysis of Electrical Machines"
     }

@inproceedings{kingma2014auto,
               author = {Diederik P {Kingma} and Max {Welling}},
               title = {Auto-Encoding Variational {Bayes}},	
               booktitle = {International Conference on Learning Representations},	
	           year = {2014}
              }

@inproceedings{kingma2015adam,
	          author = {Diederik P. {Kingma} and Jimmy Lei {Ba}},
              title = {Adam: A Method for Stochastic Optimization},
	          booktitle = {International Conference on Learning Representations},
	          year = {2015}
              }

@Article{electronics10182185,
         author = {Tucci, Mauro and Barmada, Sami and Formisano, Alessandro and Thomopulos, Dimitri},
         title = {A Regularized Procedure to Generate a Deep Learning Model for Topology Optimization of Electromagnetic Devices},
         journal = {Electronics},
         volume = {10},
         year = {2021},
         number = {18},
         article-number = {2185},
         ISSN = {2079-9292},
         doi = {10.3390/electronics10182185}
        }

@Article{app13031395,
         AUTHOR = {Raia, Maria Raluca and Ciceo, Sebastian and Chauvicourt, Fabien and Martis, Claudia},
         TITLE = {Multi-Attribute Machine Learning Model for Electrical Motors Performance Prediction},
         JOURNAL = {Applied Sciences},
         VOLUME = {13},
         YEAR = {2023},
         NUMBER = {3},
         ARTICLE-NUMBER = {1395},
         URL = {https://www.mdpi.com/2076-3417/13/3/1395},
         ISSN = {2076-3417},
         DOI = {10.3390/app13031395}
         }

@INPROCEEDINGS{8785344,
               author={Kurtović, Haris and Hahn, Ingo},
               booktitle={2019 IEEE International Electric Machines \& Drives Conference (IEMDC)}, 
               title={Neural Network Meta-Modeling and Optimization of Flux Switching Machines}, 
               year={2019},
               volume={},
               number={},
               pages={629-636},
               doi={10.1109/IEMDC.2019.8785344}
               }

@Inbook{Gerling2015-4,
        author="Gerling, Dieter",
        title="Induction Machines",
        bookTitle="Electrical Machines: Mathematical Fundamentals of Machine Topologies",
        year="2015",
        publisher="Springer Berlin Heidelberg",
        address="Berlin, Heidelberg",
        pages="135--188",
        isbn="978-3-642-17584-8",
        doi="10.1007/978-3-642-17584-8_4",
        url="https://doi.org/10.1007/978-3-642-17584-8_4"
        }

@book{boldea2018induction,
      title={The induction machines design handbook(2nd ed.)},
      author={Boldea, Ion and Nasar, Syed A},
      year={2018},
      publisher={CRC press},
      url = "https://doi.org/10.1201/9781315222592"
     }

@article{geuzaine2009gmsh,
         title={Gmsh: A 3-D finite element mesh generator with built-in pre-and post-processing facilities},
         author={Geuzaine, Christophe and Remacle, Jean-Fran{\c{c}}ois},
         journal={International journal for numerical methods in engineering},
         volume={79},
         number={11},
         pages={1309--1331},
         year={2009},
         publisher={Wiley Online Library}
        }

@article{rumelhart1986learning,
	author = {Rumelhart, David E. and Hinton, Geoffrey E. and Williams, Ronald J.},
	doi = {10.1038/323533a0},
	journal = {nature},
	number = {6088},
	pages = {533--536},
	publisher = {Nature Publishing Group},
	title = {Learning representations by back-propagating errors},
	volume = {323},
	year = {1986}}

@article{mckay2000comparison,
        author = {Mckay, M.D. and Beckkman, R.J. and Conover, William},
        year = {2000},
        month = {01},
        pages = {266-294},
        title = {Comparison of three methods for selecting values of input variables in the analysis of output from a computer code},
        volume = {21},
        journal = {Technometrics},
        doi = {10.1080/00401706.2000.10485979}
        }

@online{amini2021mit,
  author = {Amini, Alexander and Amini, Ava},
  title = {MIT 6.S191: Introduction to Deep Learning},
  year = {2021},
  url = {https://introtodeeplearning.com/},
  note = {Lecture note},
  organization = {Massachusetts Institute of Technology}
}

@ARTICLE{9745918,
  author={Di Barba, Paolo},
  journal={IEEE Transactions on Magnetics}, 
  title={Future Trends in Optimal Design in Electromagnetics}, 
  year={2022},
  volume={58},
  number={9},
  pages={1-4},
  doi={10.1109/TMAG.2022.3164204}
  }

@ARTICLE{10036443,
  author={Sato, Hayaho and Igarashi, Hajime},
  journal={IEEE Transactions on Magnetics}, 
  title={Fast Topology Optimization for PM Motors Using Variational Autoencoder and Neural Networks With Dropout}, 
  year={2023},
  volume={59},
  number={5},
  pages={1-4},
  doi={10.1109/TMAG.2023.3242288}}

@INPROCEEDINGS{9773565,
  author={Shim, Jaehoon and Lim, Gyu Cheol and Ha, Jung-Ik},
  booktitle={2022 IEEE Applied Power Electronics Conference and Exposition (APEC)},
  title={Unsupervised Anomaly Detection for Electric Drives Based on Variational Auto-Encoder}, 
  year={2022},
  volume={},
  number={},
  pages={1703-1708},
  doi={10.1109/APEC43599.2022.9773565}}

@ARTICLE{9896140,
  author={Shimizu, Yuki and Morimoto, Shigeo and Sanada, Masayuki and Inoue, Yukinori},
  journal={IEEE Transactions on Energy Conversion}, 
  title={Automatic Design System With Generative Adversarial Network and Convolutional Neural Network for Optimization Design of Interior Permanent Magnet Synchronous Motor}, 
  year={2023},
  volume={38},
  number={1},
  pages={724-734},
  doi={10.1109/TEC.2022.3208129}}

@book{Goodfellow-et-al-2016,
    title={Deep Learning},
    author={Ian Goodfellow and Yoshua Bengio and Aaron Courville},
    publisher={MIT Press},
    note={\url{http://www.deeplearningbook.org}},
    year={2016}
}
